\pdfoutput=1

\documentclass[11pt]{article}

\usepackage[preprint]{acl}

\usepackage{times}
\usepackage{latexsym}

\usepackage[T1]{fontenc}

\usepackage[utf8]{inputenc}

\usepackage{microtype}

\usepackage{inconsolata}

\usepackage{graphicx}

\usepackage{booktabs}
\usepackage{multirow}
\usepackage{makecell}
\usepackage{subcaption}
\usepackage{todonotes}
\usepackage{xspace}
\usepackage[inline]{enumitem}
\usepackage{amsmath}

\newcommand{\rescell}[2]{\makecell{$#1{\scriptstyle \pm#2}$}}

\newcommand{\narrowbotc}[1]{{\colorbox{yellow}{\parbox[t][][t]{23em}{#1}}}}
\newcommand{\liahr}{\textit{\mbox{LiaHR}}\xspace}
\newcommand{\filterbert}{\textit{\mbox{FilterBERT}}\xspace}

\title{Humans Hallucinate Too: Language Models Identify and Correct\\ Subjective Annotation Errors With Label-in-a-Haystack Prompts}

\author{
 \textbf{Georgios Chochlakis},
 \textbf{Peter Wu},
 \textbf{Arjun Bedi},\\
 \textbf{Marcus Ma},
 \textbf{Kristina Lerman},
 \textbf{Shrikanth Narayanan}
\\
 University of Southern California
\\
 \small{
   \textbf{Correspondence:} \href{mailto:chochlak@usc.edu}{chochlak@usc.edu}
 }
}

\begin{document}
\maketitle
\begin{abstract}
Modeling complex subjective tasks in Natural Language Processing, such as recognizing emotion and morality, is considerably challenging due to significant variation in human annotations. This variation often reflects \textit{reasonable} differences in semantic interpretations rather than mere noise, necessitating methods to distinguish between legitimate subjectivity and error.
We address this challenge by exploring \textit{label verification} in these contexts using Large Language Models (LLMs). First, we propose a simple In-Context Learning binary filtering baseline that estimates the \textit{reasonableness} of a document-label pair. We then introduce the \textit{Label-in-a-Haystack} setting: the query and its label(s) are included in the demonstrations shown to LLMs, which are prompted to predict the label(s) again, while receiving task-specific instructions (e.g., emotion recognition) rather than label copying.
We show how the failure to copy the label(s) to the output of the LLM are task-relevant and informative. Building on this, we propose the \textbf{L}abel-\textbf{i}n-\textbf{a}-\textbf{H}aystack \textbf{R}ectification (\liahr) framework for subjective label correction: when the model outputs diverge from the reference gold labels, we assign the generated labels to the example instead of discarding it. This approach can be integrated into annotation pipelines to enhance signal-to-noise ratios. Comprehensive analyses, human evaluations, and ecological validity studies verify the utility of \liahr for label correction. Code is available at \url{https://github.com/gchochla/liahr}.
\end{abstract}

\section{Introduction}

\begin{figure}[t]
    \centering
    \includegraphics[width=1\linewidth]{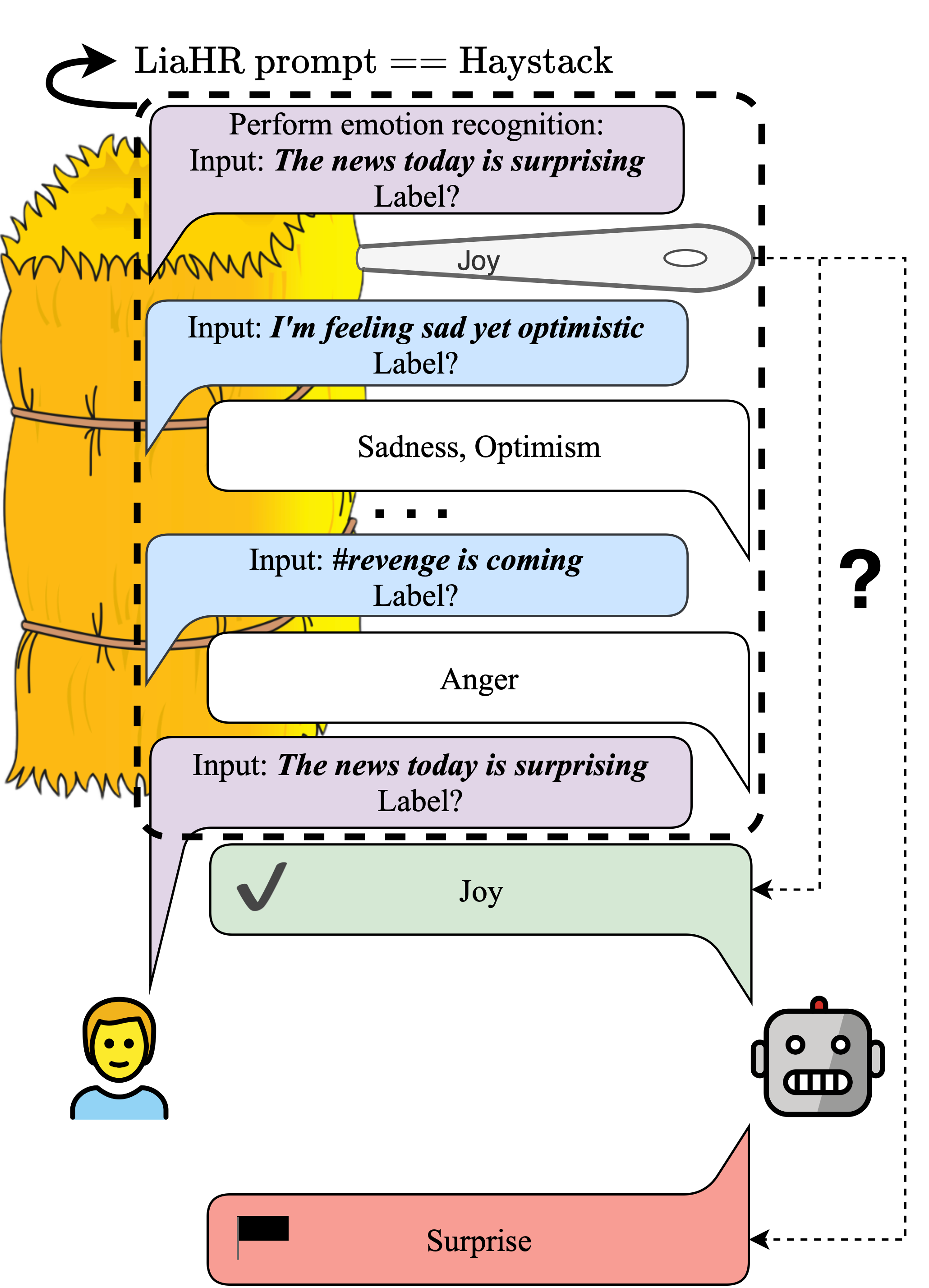}
    \caption{\textit{Label-in-a-Haystack Rectification (\textbf{LiaHR})}: The query also appears in the prompt as a demo. The LLM is instructed to perform the actual task, as captured by the label names. We leverage the failure to correctly copy-paste the query's label to flag the query-label pair, for filtering or even correction based on the prediction.}
    \label{fig:copy-paste-task}
\end{figure}

In this work, we address the challenge of modeling complex subjective tasks in natural language, captured in benchmarks such as for emotion recognition and moral foundation prediction. By ``complex subjective'', we refer to problems where multiple (subjective) interpretations can be \textit{reasonable}, and there is often no single ``correct'' answer. In such cases, ``ground'' truth is substituted with crowd truth~\cite{aroyo2015truth}, such as majority vote. Previous work has also referred to these settings as \textit{survey settings}~\cite{resnick2021survey}, where similarly ``ground'' truth is the wisdom of the crowd. This stands in contrast to ``objective'' tasks where we can define a correct answer and annotator disagreement is generally viewed as error or noise. The distinction is evident when looking at inter-annotator agreement in these settings~\cite{mohammad2018semeval, demszky2020goemotions}, but also the utility of objectively correct responses compared to disagreements in reinforcement learning with verifiable rewards~\cite{guo2025deepseek}, for instance.

Therefore, whereas noise in objective labels needs to be discarded and can be detected by looking at agreement between annotators, for subjective tasks, annotator disagreement may carry signal rather than noise, reflecting differences in perspective or background. Therefore, conventional error correction approaches based on agreement metrics are not directly applicable. Instead, improving subjective modeling requires filtering variation due to error in gold labels while preserving meaningful disagreement~\cite{booth2024people}.

To address this challenge, we propose a framework that uses LLMs for \textit{error detection and correction} in subjective annotations that respects different perspectives. In this manner, we can maintain the diversity of opinions in the data, while also maximizing the signal-to-noise ratio. Prior works in these settings~\cite{hovy2013learning, swayamdipta2020dataset, mokhberian2022noise} have relied on training classifiers across entire datasets to identify unreliable labels based on model predictions and inter-annotator disagreement. In contrast, our approach leverages LLMs in a few-shot, online setting to assess and even refine labels during annotation. We begin by introducing ``reasonableness'' labels as the simple baseline (Figure~\ref{fig:reasonableness-labels} in the Appendix) to demonstrate how LLMs can be catered to filtering explicitly instead of proxy filtering through classification. This binary indicator characterizes whether a document-label pair is reasonable (i.e., plausible, as we do not necessarily adopt a right-wrong split). We can, thereafter, prompt an LLM to predict the reasonableness label of a query document-label pair.

To achieve correction, we introduce the \textit{Label-in-a-Haystack} task, shown in Figure~\ref{fig:copy-paste-task}, that leverages the biases of LLMs toward their prior knowledge~\cite{shi2024trusting, chochlakis2024strong}. In this setting, the query and a candidate label are included in the prompt, and the model is instructed to perform the task of the dataset (that is, not merely to copy the label). Given the prediction of the LLM, we simply check whether the model was able to copy the labels from its prompt correctly. We refer to this setting as \textbf{L}abel-\textbf{i}n-\textbf{a}-\textbf{H}aystack \textbf{R}ectification (\liahr), as the model generates alternatives when it ``disagrees'' enough with the provided labels, effectively correcting unreasonable annotations.


To evaluate our proposed approaches, we first propose, define and evaluate four proxy properties integral to subjective modeling: \textbf{Nonconformity}, \textbf{Diversity}, \textbf{Noise rejection}, and \textbf{Rectification}. Then, we verify whether model decisions and proposed alternatives align well with human judgments. Finally, to assess the ecological validity of the filtering and correction proposed, we show that the performance of BERT-based models~\cite{devlin2019bert} increases on the corrected datasets.

Our findings reveal that both the reasonableness baseline and the \liahr framework can successfully verify and correct subjective labels. As such, our proposed framework can be effectively used \textit{during} (not after) the annotation process, and is specifically catered to complex subjective settings. We leverage its commonsense priors to correct the labels, rejecting unreasonable annotations in context, reinforcing prior observations that in-context learning in LLMs may rely more on \textit{task recognition} than \textit{task learning}~\cite{min2022rethinking}. Furthermore, by causally manipulating the prompt labels to belong to in-group or out-group members~\cite{dorn2024harmful}, but without explicit mention of this manipulation to the model, we show how \liahr can reliably pick up implicit cues from a few examples. Finally, we also show that aggregated labels are rejected at higher rates compared to individual annotators, corroborating previous findings~\cite{chochlakis2025aggregation} of the unsuitability of aggregation for subjective language tasks.

\section{Related Work}

\subsection{Viewpoint Diversity}

Many works have attempted to model individual annotator perspectives instead of the aggregate to capture their differing perspectives. Recently, \citet{gordonJuryLearningIntegrating2022} fused Transformer features~\cite{vaswaniAttentionAllYou2017} with annotator features that include demographic factors, among others, to model individual perspectives. Demographic information has also been fused into word embeddings by~\citet{gartenIncorporatingDemographicEmbeddings2019}. In addition, systematic biases have been assessed through rigorous annotations and profiling~\cite{sapAnnotatorsAttitudesHow2022}. Other recent work has tried to model annotators on top of common representations~\cite{davaniDealingDisagreementsLooking2022, mokhberian2023capturing}, and to decrease annotation costs online based on disagreement~\cite{golazizian2024cost}. Modeling annotators with LLMs has shown limited success due to LLM biases~\cite{dutta2023modeling, abdurahman2024perils, hartmann2023political}.

\subsection{Error Detection}

Previously, error detection has been carried out in a variety of ways and levels of intervention.
One research thread assumes a single correct answer per item, and proceed to identify errors or ``spammer'' annotators. Examples include the Dawid-Skene algorithm~\cite{dawidMaximumLikelihoodEstimation1979a}, MACE~\cite{hovy2013learning}, and CrowdKit~\cite{ustalov2021learning}. However, these methods fail the basic assumption of our work, as they do allow difference in opinion, marginalizing idiosyncratic viewpoints, which may otherwise be internally consistent~\cite{chochlakis2025aggregation}. Similar approaches that allow for disagreement still assign scores per item and annotator individually and not for separately for each pair, like CrowdTruth~\cite{CrowdTruth2}.

In another research thread, again as a post-processing step, previous work has used trained models on the dataset to assess the quality of the labels, either directly, e.g., with dataset cartography~\cite{swayamdipta2020dataset, mokhberian2022noise}, where each data point is mapped onto a 2D space depending on the confidence and the accuracy of the predictions, or indirectly, e.g., with self distillation~\cite{stanton2021does}.

Label verification has also been explored online by using predictions from a model, such as a Large Language Model (LLM), and checking them against the annotations~\cite{feng2024foundation}. However, this method trivially considers differing perspectives invalid. Previous work has also shown how the prior biases of LLMs ossify their posterior predictions~\cite{chochlakis2024strong}, which in turn leads to failures in accommodating different perspectives during regular inference. This further narrows the breadth of subjective assessment we ideally want to capture and limits our ability to potentially elicit different predictions from LLMs in subjective settings. When iterating in batches, verification checks cannot be automated similar to the aforementioned post-processing step due to the lack of sufficient data, so checks need to be manual, such as analyzing disagreement or having annotators engage in consensus talks~\cite{paletz2023social}, significantly increasing costs. \citet{liu2023we} showed that LLMs do not model ambiguity, an important component of disagreement.

\section{Methodology}

First, it is important to provide some working definition of \textit{reasonableness} (and in turn, what a subjective task is). For our purposes, we consider a document-label pair to be reasonable if and only if a person who would annotate differently can nonetheless consider some reasoning process that leads to that label valid. That is, if a human can agree that a reasoning process is \textit{valid}, \textit{coherent}, and \textit{faithful}~\cite{jacovi2020towards} with respect to the label, then that label is deemed reasonable\footnote{in the case of initial disagreement with a specific rationale, iterative refinement until agreement is achieved is valid, assuming the reasoning remains faithful to the labels}. 
We present the general and intuitive description of our methods in this section and a more mathematically rigorous description in Appendix~\ref{sec:appendix-method}.

\paragraph{Reasonableness labels} We construct a dataset dynamically, wherein our data consist of document-label pairs. As a proxy for reasonableness, the labels are either the gold label of the document from the original dataset, or randomly sampled for unreasonable pairs. This setting is shown in Figure~\ref{fig:reasonableness-labels} in the Appendix. Each document can appear with both types of labels. We sample the labels of another example from the dataset for unreasonable pairs to maintain the label distribution.

\paragraph{Label-in-a-Haystack} As shown in Figure~\ref{fig:copy-paste-task}, the query and its candidate label are included in the prompt as the first example, and the model is instructed to perform the task described by its labels. However, due to inclusion of the label in the prompt already, we essentially check whether the model is able to copy-paste the query's label onto its output. Given previous results about the collapse of the posterior predictions of LLMs to the priors in complex subjective tasks~\cite{chochlakis2024strong}, we expect that in cases where the gold labels are ``judged'' to be erroneous by the model, the copy-pasting will fail, flagging a label for further review. In addition to this ability, this setting also allows us to immediately get alternative labels for the example, a property that the baseline does not possess. In this manner, we do not waste data by discarding examples that are flagged by the model. We note that when using random labels for the query document, we sample them from a random document in the dataset, similar to the baseline.

Intuitively, this method exploits the reliance of the model on its prior knowledge of the task. If a label has sufficiently ``high probability'' for a model a priori, even if not its dominant prediction, then we expect its presence in the prompt to ``push'' the posterior towards that label enough so that it prevails in the output. Therefore, only highly unreasonable labels are rejected by the model, leading to higher precision in identifying errors. Note that the performance of the model using In-Context Learning is rather poor for such tasks~\cite{chochlakis2024strong}, resulting in poor precision with many false negatives and therefore increased annotation costs. 




\subsection{Proxy Properties}

In this section, we define and present desirable proxy properties that can be used as proxies for the label filtering and correction ability that practitioners can use to guide model selection. Note that since strictness is not required because of their proximate nature, some of them are fuzzy and heuristic.

\begin{quote}
    \textbf{Nonconformity}: The model should flag some dataset labels as unreasonable, but only for a small percentage of examples.
\end{quote}

This is the first requirement. Although ``small'' is nebulous, the model should be copying the gold labels significantly better compared to its performance as a classifier. Having a smaller gap to the dataset's labels indicates an ability to agree with different perspectives, and it assumes that most of the dataset has been annotated properly\footnote{in this assumption, we take for granted that annotators have been screened, trained, attention-checked, etc. Namely, we assume quality data collection}.

\begin{quote}
    \textbf{Diversity}: The model should accept different labels consistently.
\end{quote}

Respecting different opinions is also an integral property. Here, we also assume that most annotators have annotated most of the dataset properly\footnote{we make the same assumptions as above}.
For this quality of the model, we can use annotations from different individuals and expect the model to predict reasonableness or successfully copy the labels at equally high rates for them all.

\begin{quote}
    \textbf{Noise Rejection}: The model should assign reasonableness at random performance levels when using random labels.
\end{quote}

That is to say, when asked to ``verify'' a random label, the model should succeed only when the label ``happens'' to be reasonable, meaning random levels of performance (though not exactly a random baseline, as more perspectives not present in the data could also be valid). We measure this by randomizing the label of the pair for the baseline or the query label for \liahr, and expect low success rates of filtering or copying respectively.

\begin{quote}
    \textbf{Rectification}: When \liahr is prompted with random labels for the query, its alternative predictions should be closer to the original, gold labels than the random labels it was given.
\end{quote}
This final property is a \liahr-specific constraint. If the model is to not only identify unreasonable labels, but also correct them, then when it is given random labels for the query, its predictions should be closer to the gold labels compared to those random labels. As a result, this can be measured by calculating the similarity of the \liahr predictions when \liahr is provided a random query label with the original, gold labels, and comparing that with the copy performance for the random labels, which is equivalent to the similarity of the predictions to the random labels. We expect successful models to have the higher similarity to the gold labels. We expect that priming the model with random labels may cause it to fail to meet this precisely, so only trends towards it are sought out.

\subsection{Human Evaluations}

To validate our findings on the proxy properties, we perform human evaluations in two settings:
\paragraph{\textbf{Reasonableness}} We compare human assessment of the reasonableness of the labels to the LLM's assessments. We use the chi-square test of independence of variables in a contingency table to evaluate the significance of our results (with the binary variable being reasonableness).
\paragraph{\textbf{Preference}} We compare human preference for \liahr predictions over the gold label. Significance is calculated with a binomial test.
We also compare to the regular ICL predictions to isolate the effects of \liahr from the model's classification capabilities on these tasks.

\subsection{Ecological Validity} \label{sec:eco-val-method}

In addition to human evaluations, we train smaller models on the labels derived from our filtering pipelines. Namely, we examine
\begin{enumerate*}[label=(\roman*)]
    \item the \texttt{Original} labels,
    \item \liahr on the entire corpus (\texttt{Replaced}) or only on the \texttt{(trn)} set,
    \item \liahr but used to filter training example when copy-pasting is erroneous (\texttt{Filtered}),
    \item the \textit{reasonableness} baseline to filter out training examples (\texttt{Bsl Filtered}),
    \item the \texttt{Predictions} of the LLM with ICL.
\end{enumerate*}

\subsection{Metrics}

Because the \liahr format is identical to classification, we use classification metrics to evaluate the performance of copy-pasting and get a more nuanced picture of the predictions of the model. We use Jaccard Score and Micro F1 for multilabel, and accuracy and F1 for single-label cases.

\section{Experiments}

\subsection{Datasets}

\paragraph{SemEval 2018 Task 1 E-c (SemEval; \citealt{mohammad2018semeval})} A multilabel emotion recognition benchmark containing annotations for 11 emotions: \textit{anger}, \textit{anticipation}, \textit{disgust}, \textit{fear}, \textit{joy}, \textit{love}, \textit{optimism}, \textit{pessimism}, \textit{sadness}, \textit{surprise}, and \textit{trust}. We use only the English subset.

\begin{figure}[t]
    \centering
    \includegraphics[width=1\linewidth]{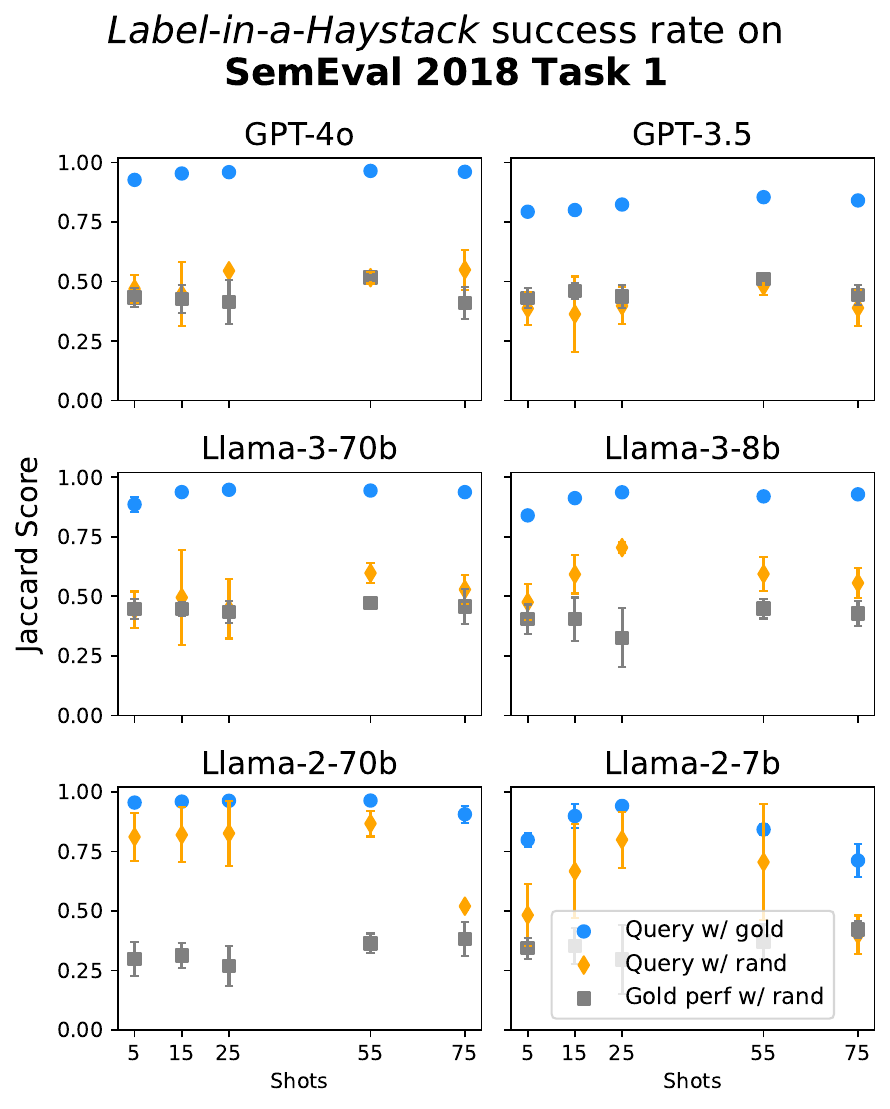}
    \caption{Success rate of copying the labels in \liahr on \textbf{SemEval} when using the gold and random labels for the query in the prompt across various numbers of demonstrations. We also show performance w.r.t. the gold labels when using random query labels.}
    \label{fig:semeval-copy-paste-results}
\end{figure}

\begin{figure}[t]
    \centering
    \includegraphics[width=1\linewidth]{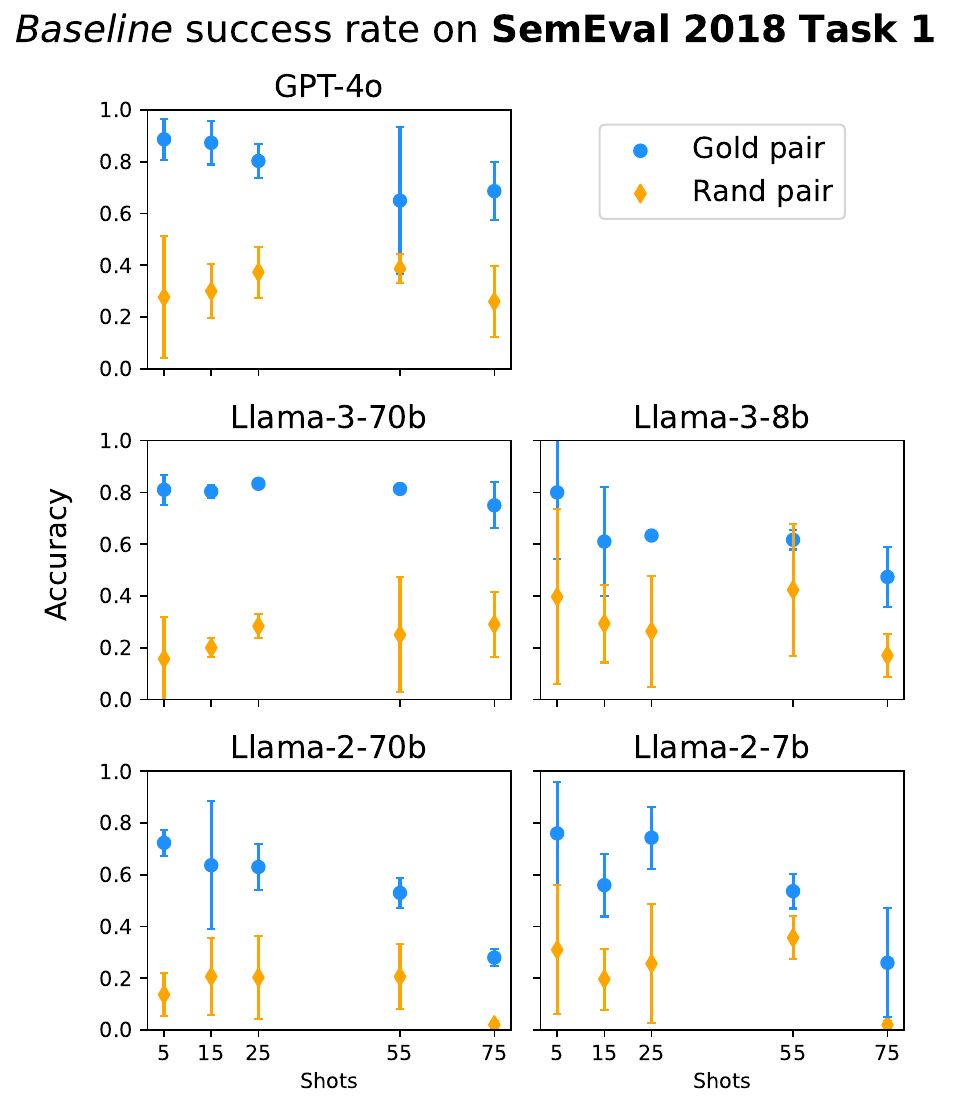}
    \caption{\textit{Baseline} ``reasonable'' scores on \textbf{SemEval} when using gold and random input-label pairs.}
    \label{fig:semeval-baseline-results}
\end{figure}

\paragraph{MFRC~\cite{trager2022moral}} A multilabel moral foundation corpus with annotations for six moral foundations: \textit{care}, \textit{equality}, \textit{proportionality}, \textit{loyalty}, \textit{authority}, and \textit{purity}. The dataset was released with annotator labels.

\paragraph{GoEmotions~\cite{demszky2020goemotions}} A multilabel emotion recognition benchmark with 27 emotions. For efficiency and conciseness, we pool the emotions to the following seven ``clusters'' using hierarchical clustering: \textit{admiration}, \textit{anger}, \textit{fear}, \textit{joy}, \textit{optimism}, \textit{sadness}, and \textit{surprise}. The dataset was released with annotator labels.

\paragraph{QueerReclaimLex~\cite{dorn2024harmful}} Single-label binary harm dataset, which contains various templates populated with reclaimed LGTBQ+ slurs. It contains two harm labels: assuming in-group and out-group authors. Using one or the other without explicit mention, we can evaluate the \textbf{Diversity} property with a known and controllable causal factor. This setting serves as a stress test, since reclaimed slurs in general are a documented failure case for, e.g., toxicity classifiers~\cite{sap2019risk, haimson2021disproportionate, sapAnnotatorsAttitudesHow2022}, allowing us to examine whether systematic biases in LLMs influence their decisions in our framework. For the same reasons, it is challenging because it includes a realistic confounding factor: the interplay between politeness guardrails and our desired behavior, as slurs are explicitly included throughout the prompt. We create splits to be as balanced as possible, but also present ROC-AUC to avoid bias. Because the labels are binary, we use the opposite label instead of randomizing the query label.

\begin{figure}[t]
    \centering
    \includegraphics[width=1\linewidth]{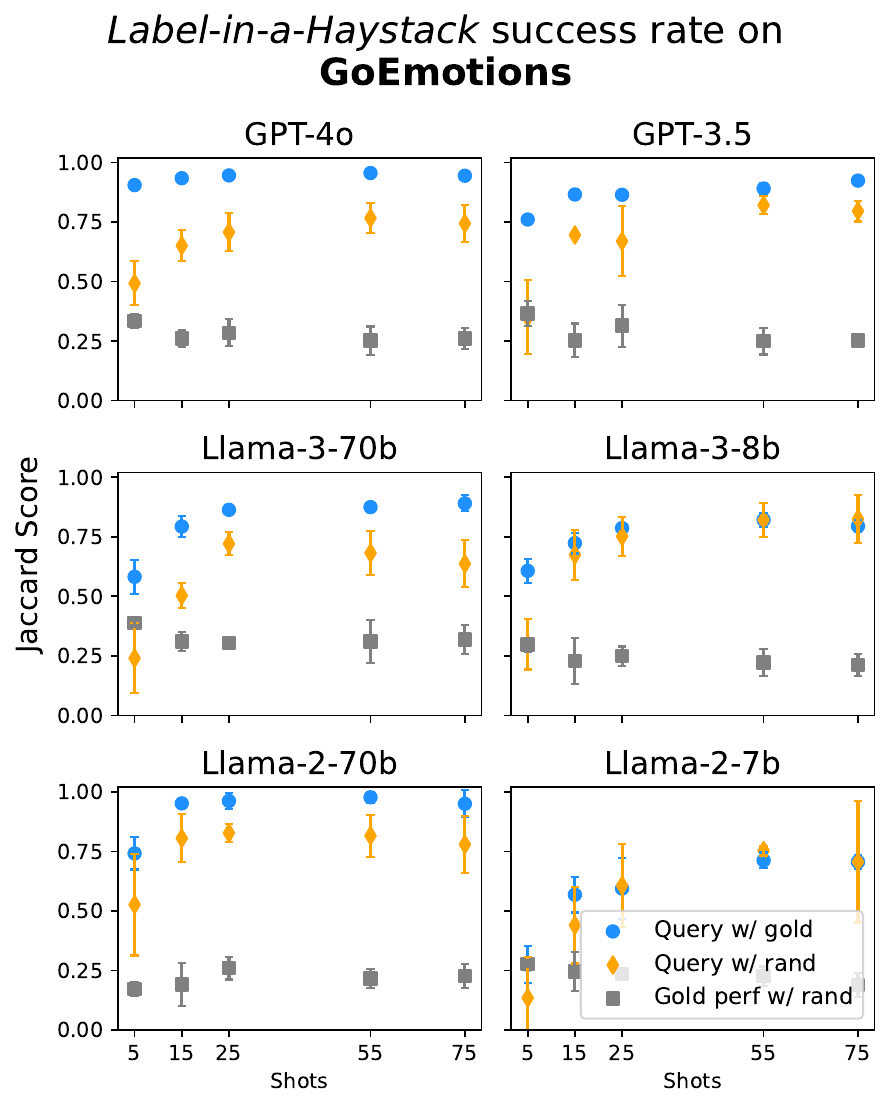}
    \caption{Success rate of copying the labels in \liahr on \textbf{GoEmotions} when using the gold and random labels for the query in the prompt across various numbers of demonstrations. We also show performance w.r.t. the gold labels when using random query labels.}
    \label{fig:goemotions-copy-paste-results}
\end{figure}

\begin{figure}[t]
    \centering
    \includegraphics[width=1\linewidth]{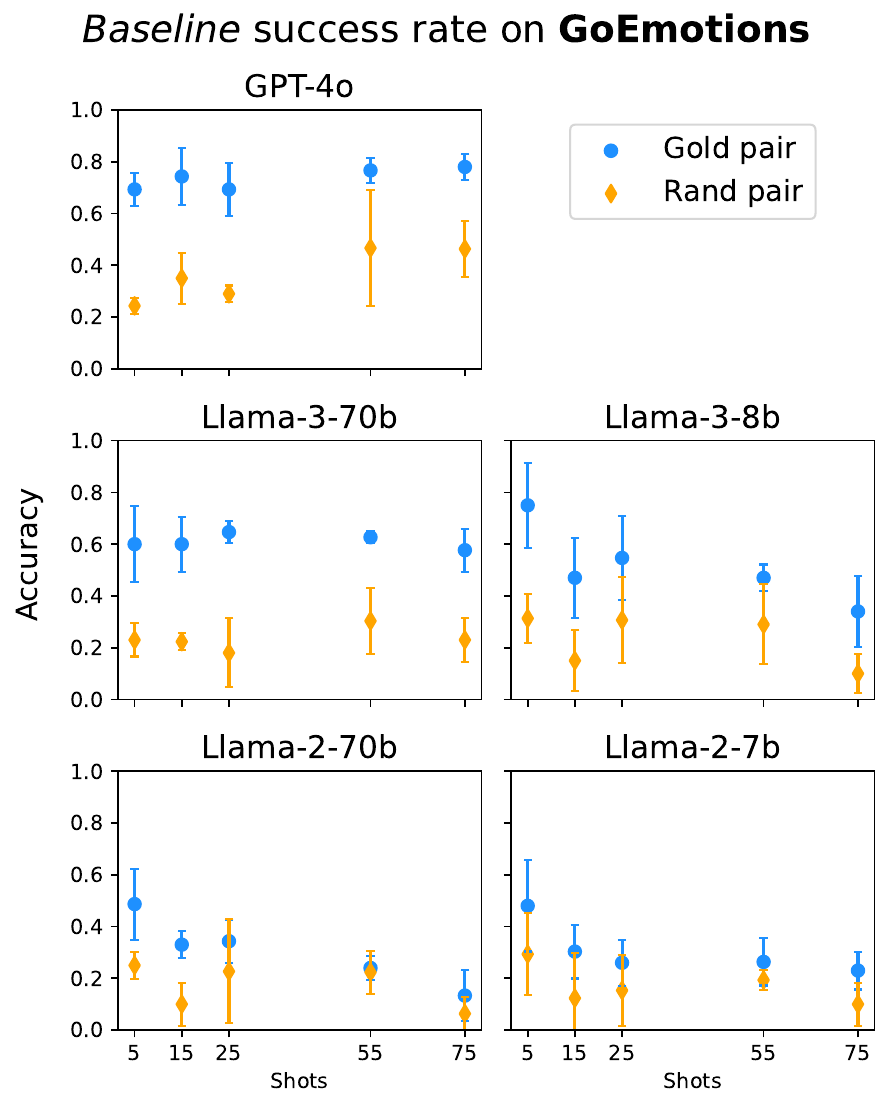}
    \caption{\textit{Baseline} ``reasonable'' scores on \textbf{GoEmotions} when using gold and random input-label pairs.}
    \label{fig:goemotions-baseline-results}
\end{figure}

\subsection{Implementation Details}

We use the 4-bit quantized versions of the open-source LLMs through the \textit{HuggingFace}~\cite{wolf-etal-2020-transformers} and \texttt{bitandbytes} interface for \textit{PyTorch}. We use GPT-3.5 Turbo (\texttt{gpt-3.5-turbo}), GPT-4 (\texttt{gpt-4-turbo}), and GPT-4o (\texttt{gpt-4o-mini}), Llama-2 7B and 70B (\texttt{meta-llama/Llama-2-\#b-chat-hf}), and Llama-3 8B and 70B (\texttt{meta-llama/Meta-Llama-3-\#B-Instruct}). We chose only finetuned models~\cite{ouyang2022training} to avoid confounding factors. We use random retrieval of examples. We train \textit{Demux}~\cite{chochlakisLeveragingLabelCorrelations2023} as the smaller model for ecological validity. When sampling random labels, we ensure at least one label is present (i.e., we do not allow \texttt{None}s because of their higher plausibility). Results for proxy properties are 3 different seeds with 100 inference examples each. The entire corpus is used for training and evaluation of smaller models. Unless otherwise noted, we show 95\% confidence interval around the mean. For more details, see Appendix~\ref{sec:appendix-more-impl} and \ref{sec:appendix-human-eval}.

\subsection{Evaluating Proxy Properties}


The first step to applying these methods for label verification is to show that copy-pasting can fail in \liahr, and that they indeed meet the desired proxy properties. Throughout this section, when presenting \textbf{success rates}, that refers to the amount of copy-pasting that happened successfully. This means that \textit{when randomizing} the labels, we still count \textit{whether the random labels are generated}, and therefore \textit{lower scores on random labels} represent more desirable behavior.

\paragraph{SemEval} We present our results for all\footnote{some API-based models were deprecated during the course of our experiments, so we skip them where they are not available. For additional results, such as GPT-4, see Appendix~\ref{sec:appendix-deprecated-semeval}.} models in Figure~\ref{fig:semeval-copy-paste-results} for \liahr and Figure~\ref{fig:semeval-baseline-results} for the baseline. In Figure~\ref{fig:semeval-copy-paste-results}, we present the performance of the model on the copy-paste task when using gold (\texttt{Query w/ gold}) and random (\texttt{Query w/ rand}) labels for the demo query, as well as the performance of the model on the gold labels when the query label is random (and therefore the model has not seen the test label for the query; \texttt{Gold perf w/ rand}). All results are shown for 5, 15, 25, 55, and 75 shot (to demonstrate scalability). For Figure~\ref{fig:semeval-baseline-results}, we show the first two scenarios, where the document is presented to the LLM with its paired label (\texttt{Gold pair}) or a random label (\texttt{Rand pair}).

In \liahr, we see clear evidence for our desired behavior in bigger and more capable models, specifically GPT-3.5, GPT-4o, and Llama-3 70b. These models seem to display all the properties we check for: \textbf{Nonconformity}, \textbf{Rectification}, and \textbf{Noise rejection}. First, the success rate with gold labels for the query is not perfect (meaning 1.0), yet it is significantly higher compared to the same model's performance on the benchmark~\cite{chochlakis2024strong}. This means that the model does not conform to the gold labels completely, yet is greatly influenced by them in its predictions (otherwise we would anticipate performance much closer to its ``regular'' predictions). By meeting both these criteria, the aforementioned models meet the \textbf{Nonconformity} property. Then, when we use random labels instead of gold for the query in the prompt, we see the success rate drop dramatically compared to when gold labels are presented (that it, when comparing \texttt{Query w/ gold} to \texttt{Query w/ rand}). This indicates that models achieve the \textbf{Noise Rejection} property. Moreover, it is interesting to see that, when random labels are provided, the predictions match more closely the gold labels (\texttt{Gold perf w/ rand}) than these random labels (\texttt{Query w/ rand}). Since this criterion is met, the models achieve \textbf{Rectification}.

For the ``reasonableness'' baseline, we see that only GPT-4o meets the criteria \textbf{Nonconformity}, and \textbf{Noise rejection}\footnote{\textbf{Rectification} is not a potential property because the LLM does not generate labels}. While other models mostly meet the \textbf{Noise Rejection} criterion, their success rate is too low to qualify for \textbf{Nonconformity}. We also notice that the success rate in all settings is noticeably lower compared to \liahr.

Interestingly, when looking at smaller and less capable models, we see that the models achieve higher copy-paste performance, both with the dataset labels and with random labels, and therefore \textbf{Nonconformity} and \textbf{Noise Rejection} are only partially achieved. Consequently, when using random query labels, their predictions are more similar to these random labels compared to the dataset labels, so the models do not display \textbf{Rectification}.

\begin{figure}[!t]
    \centering
    \includegraphics[width=0.95\linewidth]{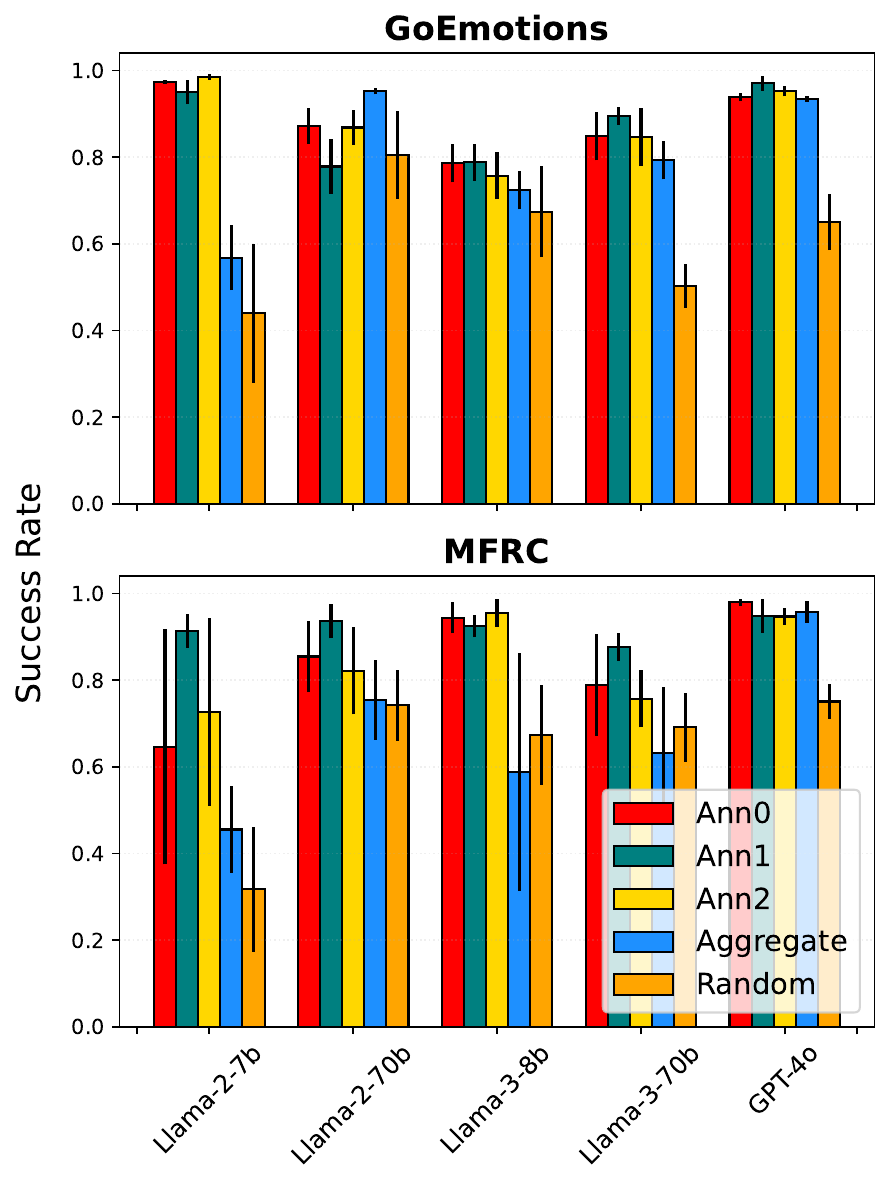}
    \caption{Success rate of copying the labels of \liahr on \textbf{GoEmotions} and \textbf{MFRC} with aggregate labels, random labels, and annotator labels (\textit{Ann\#}), shown for 15-shot prompts. For GoEmotions, actual annotator IDs are: Ann0 = 37, Ann1 = 4, Ann2 = 61. For MFRC: \mbox{Ann0 = 0}, Ann1 = 1, Ann2 = 3.}
    \label{fig:diversity}
\end{figure}

\begin{figure}[!ht]
    \centering
    \includegraphics[width=0.95\linewidth]{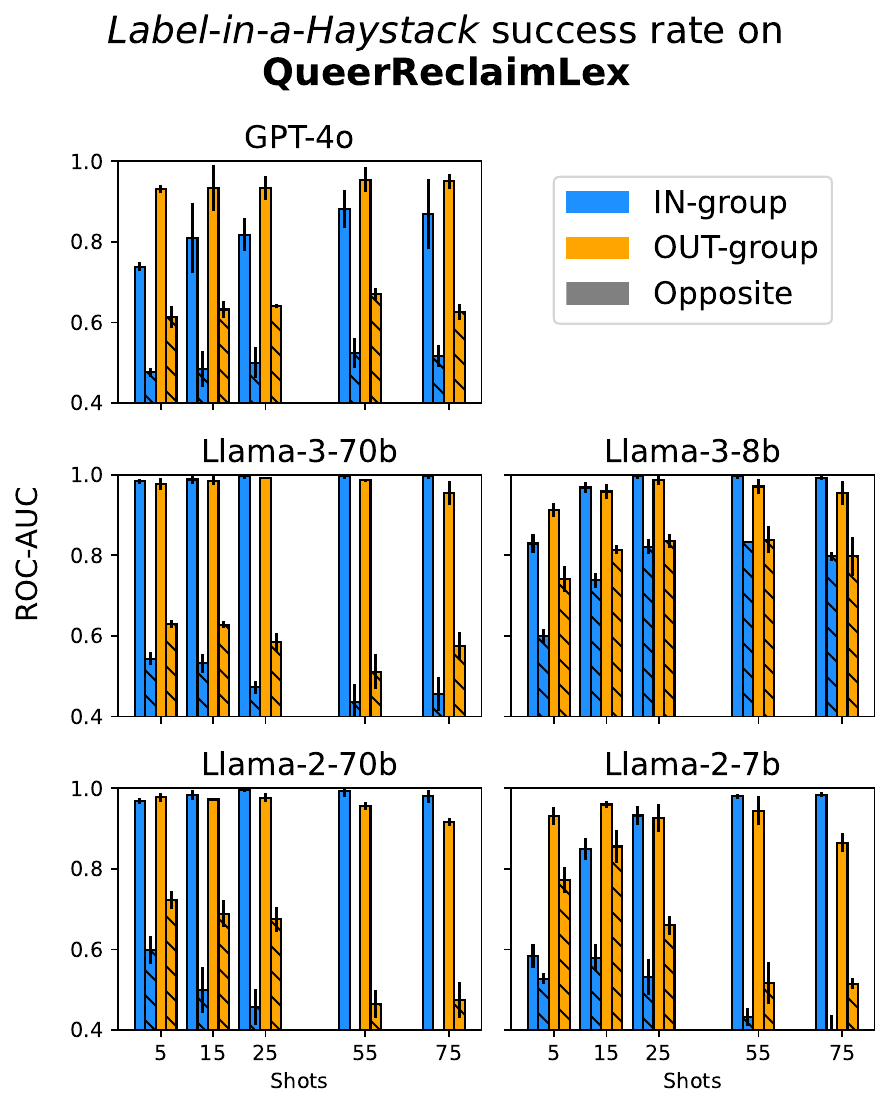}
    \caption{Success rates on copying labels in \liahr on \textbf{QueerReclaimLex} when using in-group labels or out-group labels in the prompt as a proxy for \textbf{Diversity}. Query included with current group's label or \textit{opposite}.}
    \label{fig:diversity-queer}
\end{figure}

\begin{table}[!t]
    \centering
    \begin{tabular}{l cc}
        \hspace{-8px}\liahr & \textbf{Reasonableness} & \textbf{Preference} \\
        \midrule
        Llama-3 70b & 6.57e-1 & \textbf{3.36e-2} \\
        GPT-3.5 & 9.52e-2 & 7.25e-2 \\
        GPT-4 &  \textbf{2.38e-7} & \textbf{6.86e-4} \\
        GPT-4o & \textbf{1.40e-4} & \textbf{5.08e-5} \\\\
        \hspace{-8px} \textit{Baseline} \\
        \midrule
        Llama-3 70b & \textbf{6.11e-4} & - \\
        GPT-4o & \textbf{8.08e-10} & - \\\\
        \hspace{-8px} \textit{ICL} \\
        \midrule
        GPT-3.5 & - & 5.19e-1 \\
        GPT-4 & - & 1 \\
        
    \end{tabular}
    \caption{\textit{p-values} for \liahr on \textbf{SemEval}. \textbf{Reasonableness} refers to whether human and LLM unreasonable assessments coincide. \textbf{Preference} to whether humans prefer model predictions over gold labels. \textit{p-values} are for the hypothesis that the models agree with humans.}
    \label{tab:significance}
\end{table}

\paragraph{GoEmotions} We show our results for \liahr in Figure~\ref{fig:goemotions-copy-paste-results} and for the baseline in Figure~\ref{fig:goemotions-baseline-results}. We notice that in GoEmotions, even GPT-4o struggles, with the acceptance rates of random labels, as the gap is smaller to the gold labels when compared to SemEval. Therefore, it is evident that only a small subset of the settings is able to \textit{clearly} achieve \textbf{Nonconformity} and \textbf{Noise Rejection}, namely 5-shot GPT-4o, 5-shot GPT-3.5, and 15-shot Llama-3 70b, while these models also seem to be meeting or tending towards \textbf{Rectification}. Again, the baseline, on the other hand, seems to be achieving consistently lower success rates for the gold labels, but their random performance is much lower and therefore better at \textbf{Noise Rejection}.

\paragraph{MFRC} In Appendix~\ref{sec:appendix-mfrc-properties}, we also show our results and very interesting findings for these three properties in MFRC, where smaller models seem to be treating the gold and random labels similarly.

\paragraph{BERT-based baseline} In Appendix~\ref{sec:appendix-filterbert}, we show results for a BERT-based filtering baseline, showing it underperforms 5-shot GPT-4o while requiring the entire dataset to be trained, disincentivizing its usage from beginning to end of the data collection.

\paragraph{Diversity} We examine \textbf{Diversity} separately, in Figures~\ref{fig:diversity} and \ref{fig:diversity-queer}. Figure~\ref{fig:diversity} shows the success rates of copy-pasting on MFRC and GoEmotions between the gold, random, and individual annotator labels, using otherwise the same exact prompts and only differing the labels to avoid confounding factors. We first see that all annotators tend to be clustered together with small rejection rates, indicating that the model tends to accept all different perspectives equally. Second, we see that their performance is better compared to random. Finally, the similarity between the annotators shown can be very low (e.g., as low as 0.433 Jaccard Score on GoEmotions between annotators), representing consistently different perspectives. Consequently, the \textit{majority of the disagreement between annotators is being preserved by the model} organically, without any intervention. All these pieces of evidence indicate that most models achieve \textbf{Diversity}. Moreover, we see a marked difference between annotators and the aggregate, with the latter displaying higher rejection rates, indicating that part of our aforementioned results on MFRC and GoEmotions can be explained as aggregation artifacts~\cite{chochlakis2025aggregation}.

Figure~\ref{fig:diversity-queer} shows that \liahr can successfully accept both in-group and out-group perspectives in the QueerReclaimLex benchmark without explicit prompting, instead learning implicit causal cues from few examples. Results show that the models tend to model out-group annotations better. However, more capable models also recognize reclaimed slurs as not harmful when used by in-group speakers, scaling performance with more demonstration, indicating the robustness of \liahr to the guardrails placed on models, and an ability to counterbalance systematic biases with few demonstrations, a challenging problem in toxicity classifiers with reclaimed slurs~\cite{sap2019risk, haimson2021disproportionate, sapAnnotatorsAttitudesHow2022}. Thus, \liahr proves robust to our stress test for whether the model itself might introduce biases in the dataset.

\begin{table}[!t]
    \centering
    \begin{tabular}{lcc}
        \multirow{2}{*}{\textbf{Setting}} & \multicolumn{2}{c}{\textbf{Micro F1}} \\
        \cmidrule(lr){2-3}
        & \textbf{GoEmotions} & \textbf{SemEval} \\
        \midrule
        \texttt{Original} & \rescell{0.652}{0.001} & \rescell{0.689}{0.002} \\
        \texttt{Replaced} & \rescell{\textbf{0.653}}{0.000} & \rescell{\textbf{0.692}}{0.003} \\
        \texttt{Replaced (trn)} & \rescell{0.642}{0.001} & \rescell{0.680}{0.002} \\
        \texttt{Filtered} & \rescell{0.652}{0.002} & \rescell{0.679}{0.002} \\
        \texttt{Bsl Filtered} & \rescell{0.638}{0.001} & \rescell{0.680}{0.003} \\
        \texttt{Predictions} & \rescell{0.427}{0.002} & \rescell{0.613}{0.000} \\
    \end{tabular}
    \caption{Performance of BERT-based \textit{Demux} on various settings using \liahr and baseline label corrections.}
    \label{tab:microf1-setting}
\end{table}

\subsection{Human Evaluation}

Results for our human evaluations are presented in Table~\ref{tab:significance} for SemEval for the models that meet our defined properties. More detailed results on \textbf{SemEval} and \textbf{GoEmotions} can be found in Appendix~\ref{sec:appendix-human-eval}. We see that Llama-3 70b and GPT-3.5 do not show enough discriminability between reasonable and unreasonable labels, although their results are strong in terms of preference for their labels when the copy-paste task was performed incorrectly. However, GPT-4 and 4o can distinguish between reasonable and unreasonable labels and also propose better alternatives for unreasonable labels. The results show strong statistical significance, but also large effect sizes. This is not the case when checking for the ICL prediction of the models. This shows that the predictions of LLMs are not preferred over the gold labels by humans, indicating that our settings are important to achieve proper filtering. We also see that the explicit baseline shows sufficient discriminability for both Llama-3 70b and GPT-4o.

\subsection{Ecological Validity}

In addition to the human evaluations and defining and evaluating proxy properties, we also perform ecological validity studies, and compare to other online methods. That is, even though we have shown the models have desirable properties, and people tend to prefer them over the original labels, do models trained on them perhaps show erratic behavior? For all the settings introduced in Section~\ref{sec:eco-val-method}, we show the results in Table~\ref{tab:microf1-setting} (additional results in Appendix~\ref{sec:appendix-eco-val}). The results indicate that the new labels lead to slightly better generalization performance, although the methods need to be applied throughout the annotation process to get the maximum benefit. Note that \textbf{SemEval} is a smaller dataset, leading to extra performance decreases when examples are filtered instead of corrected. Noticeably, we also see that using the raw predictions of the models leads to substantial deterioration in performance. In addition to the humans evaluations, these results indicate that our proposal for ``reasonableness'' checks rather than simply using the LLM as classifier is warranted.

\section{Conclusion}

In this work, we propose ``reasonableness'' checks to improve the signal-to-noise ratio in subjective language annotations. We leverage LLMs and introduce \liahr, which is able to both filter and correct unreasonable annotations, and a simple baseline that detects unreasonable annotations. We demonstrate that both approaches satisfy desirable proxy properties, pass human evaluations, and show ecological validity when used to train smaller models. Moreover, we show that the model can pick up on causal yet implicit cues from few examples reliably. While our experiments show that humans prefer the model's labels when it is performing correction, we advocate for usage during the annotation process, with additional checks. For example, if some submitted labels for a specific example do not pass the \liahr filter, instead of always using its alternative predictions, the same document can be shown to the annotator at a later stage to verify and potentially correct the label themselves.

To further corroborate our findings on \liahr, we also show how it performs in objective tasks in Appendix~\ref{sec:appendix-objective}, an analysis of the copy-paste performance across shots, model families and sizes in Appendix~\ref{sec:appendix-degradation}, that individual labels are uniformly affected in Appendix~\ref{sec:appendix-filter-per-label}, and
the robustness to the position of the query in Appendix~\ref{sec:appendix-position}.

\section{Limitations and Ethics}

We want to emphasize that our model is not an oracle. The model does not provide ground truth / gold labels and could be biased in other ways.

Our work entails some potential for deliberate misuse. Although we advocate for using individual perspectives as demonstrations in \liahr throughout our work, deliberate misuse might include skewing the perspectives in the prompt and using the rejection from \liahr as justification for rejecting minority labels and preventing certain valid perspectives from entering the data (i.e., gate-keeping). Therefore, we want to emphasize that \liahr assessments can only be considered valid (though \textit{not necessarily correct}) when the perspective being evaluated (the query label) coincides with the perspective in the demonstrations. The predictions of the model should not be taken into account otherwise.

Accidental misuse includes model biases seeping into the labels. We want to note that, despite the remarkable robustness of the framework on the reclaim slurs dataset, \textbf{QueerReclaimLex}, its performance on the in-group data is noticeably worse than the out-group. This indicates that there might be some bias in the decisions of the model. Moreover, assessments of harm are inherently subjective, reflecting differences in individual and cultural perceptions. The original work aggregates distinct gender identities (e.g., non-binary, transgender) under the umbrella term gender-queer and treats them as largely synonymous. While this simplification overlooks the diversity of perspectives, we follow the original work's adoption of this grouping as a pragmatic choice to support our analysis. As understandings of differing perspectives continue to evolve, future work should aim to incorporate a broader pool of annotators and explore methods for capturing the nuance and variability of perceived harm across different communities. Therefore, we urge \textit{immense} caution when the framework is used in sensitive settings.

We also decreased the number of inference queries within each seed to enable us to experiment with many models and shots. This tradeoff means that we do not have a high degree of confidence in each individual result, yet the vast number of experiments demonstrating similar trends reinforces our confidence in our general findings.

A potential confounding factor in our work is quantization. Previous work has reported significant decreases in performance from it~\cite{marchisio2024does}.  We note, first, that there is no a priori reason for the quantization to affect our results in a nonuniform way, e.g., affecting random labels more than gold labels. Quantization was chosen because of obvious computational constraints. Finally, it is plausible that even API-based models are served quantized (e.g., \texttt{mini} versions). For these reasons, we believe that quantized performance is representative of LLM performance in realistic scenarios. Moreover, this work does not aim to establish the benchmark performance of LLMs in any task, but rather to leverage their capabilities to solve a prescient problem in subjective annotations.

\section*{Acknowledgments}

This project was supported in part by funds from DARPA under contract HR001121C0168,  NSF CIVIC, and USC-Capital One Center for Responsible AI Decision Making in Finance. The authors thank Efthymios Tsaprazlis, Efthymios Georgiou, Kleanthis Avramidis and Sabyasachee Baruah for helpful comments.

\bibliography{library}

\begin{thebibliography}{43}
\providecommand{\natexlab}[1]{#1}

\bibitem[{Abdurahman et~al.(2024)Abdurahman, Atari, Karimi-Malekabadi, Xue, Trager, Park, Golazizian, Omrani, and Dehghani}]{abdurahman2024perils}
Suhaib Abdurahman, Mohammad Atari, Farzan Karimi-Malekabadi, Mona~J Xue, Jackson Trager, Peter~S Park, Preni Golazizian, Ali Omrani, and Morteza Dehghani. 2024.
\newblock Perils and opportunities in using large language models in psychological research.
\newblock \emph{PNAS nexus}, 3(7):pgae245.

\bibitem[{Aroyo and Welty(2015)}]{aroyo2015truth}
Lora Aroyo and Chris Welty. 2015.
\newblock Truth is a lie: {{Crowd}} truth and the seven myths of human annotation.
\newblock \emph{AI Magazine}, 36(1):15--24.

\bibitem[{Booth and Narayanan(2024)}]{booth2024people}
Brandon~M Booth and Shrikanth~S Narayanan. 2024.
\newblock People make mistakes: Obtaining accurate ground truth from continuous annotations of subjective constructs.
\newblock \emph{Behavior Research Methods}, 56(8):8784--8800.

\bibitem[{Chochlakis et~al.(2023)Chochlakis, Mahajan, Baruah, Burghardt, Lerman, and Narayanan}]{chochlakisLeveragingLabelCorrelations2023}
Georgios Chochlakis, Gireesh Mahajan, Sabyasachee Baruah, Keith Burghardt, Kristina Lerman, and Shrikanth Narayanan. 2023.
\newblock Leveraging label correlations in a multi-label setting: {{A}} case study in emotion.
\newblock In \emph{{{ICASSP}} 2023-2023 {{IEEE International Conference}} on {{Acoustics}}, {{Speech}} and {{Signal Processing}} ({{ICASSP}})}, pages 1--5. IEEE.

\bibitem[{Chochlakis et~al.(2024)Chochlakis, Potamianos, Lerman, and Narayanan}]{chochlakis2024strong}
Georgios Chochlakis, Alexandros Potamianos, Kristina Lerman, and Shrikanth Narayanan. 2024.
\newblock The strong pull of prior knowledge in large language models and its impact on emotion recognition.
\newblock In \emph{Proceedings of the 12th International Conference on Affective Computing and Intelligent Interaction (ACII)}. IEEE.

\bibitem[{Chochlakis et~al.(2025)Chochlakis, Potamianos, Lerman, and Narayanan}]{chochlakis2025aggregation}
Georgios Chochlakis, Alexandros Potamianos, Kristina Lerman, and Shrikanth Narayanan. 2025.
\newblock \href {https://doi.org/10.18653/v1/2025.naacl-long.284} {Aggregation artifacts in subjective tasks collapse large language models' posteriors}.
\newblock In \emph{Proceedings of the 2025 Conference of the Nations of the Americas Chapter of the Association for Computational Linguistics: Human Language Technologies (Volume 1: Long Papers)}, pages 5513--5528, Albuquerque, New Mexico. Association for Computational Linguistics.

\bibitem[{Davani et~al.(2022)Davani, Díaz, and Prabhakaran}]{davaniDealingDisagreementsLooking2022}
Aida~Mostafazadeh Davani, Mark Díaz, and Vinodkumar Prabhakaran. 2022.
\newblock Dealing with disagreements: {{Looking}} beyond the majority vote in subjective annotations.
\newblock \emph{Transactions of the Association for Computational Linguistics}, 10:92--110.

\bibitem[{Dawid and Skene(1979)}]{dawidMaximumLikelihoodEstimation1979a}
Alexander~Philip Dawid and Allan~M Skene. 1979.
\newblock Maximum likelihood estimation of observer error-rates using the {{EM}} algorithm.
\newblock \emph{Journal of the Royal Statistical Society: Series C (Applied Statistics)}, 28(1):20--28.

\bibitem[{Demszky et~al.(2020)Demszky, Movshovitz-Attias, Ko, Cowen, Nemade, and Ravi}]{demszky2020goemotions}
Dorottya Demszky, Dana Movshovitz-Attias, Jeongwoo Ko, Alan Cowen, Gaurav Nemade, and Sujith Ravi. 2020.
\newblock {{GoEmotions}}: A dataset of fine-grained emotions.
\newblock In \emph{Proceedings of the 58th Annual Meeting of the Association for Computational Linguistics}, pages 4040--4054.

\bibitem[{Devlin et~al.(2019)Devlin, Chang, Lee, and Toutanova}]{devlin2019bert}
Jacob Devlin, Ming-Wei Chang, Kenton Lee, and Kristina Toutanova. 2019.
\newblock Bert: Pre-training of deep bidirectional transformers for language understanding.
\newblock In \emph{Proceedings of the 2019 conference of the North American chapter of the association for computational linguistics: human language technologies, volume 1 (long and short papers)}, pages 4171--4186.

\bibitem[{Dorn et~al.(2024)Dorn, Kezar, Morstatter, and Lerman}]{dorn2024harmful}
Rebecca Dorn, Lee Kezar, Fred Morstatter, and Kristina Lerman. 2024.
\newblock Harmful speech detection by language models exhibits gender-queer dialect bias.
\newblock In \emph{Proceedings of the 4th ACM Conference on Equity and Access in Algorithms, Mechanisms, and Optimization}, pages 1--12.

\bibitem[{Dumitrache et~al.(2018)Dumitrache, Inel, Aroyo, Timmermans, and Welty}]{CrowdTruth2}
Anca Dumitrache, Oana Inel, Lora Aroyo, Benjamin Timmermans, and Chris Welty. 2018.
\newblock \href {https://arxiv.org/abs/1808.06080} {Crowdtruth 2.0: Quality metrics for crowdsourcing with disagreement}.

\bibitem[{Dutta et~al.(2023)Dutta, Mittal, Chen, Ramachandran, Rajakumar, Kivlichan, Mak, Butryna, and Paritosh}]{dutta2023modeling}
Senjuti Dutta, Sid Mittal, Sherol Chen, Deepak Ramachandran, Ravi Rajakumar, Ian Kivlichan, Sunny Mak, Alena Butryna, and Praveen Paritosh. 2023.
\newblock Modeling subjectivity (by mimicking annotator annotation) in toxic comment identification across diverse communities.
\newblock \emph{arXiv preprint arXiv:2311.00203}.

\bibitem[{Feng and Narayanan(2024)}]{feng2024foundation}
Tiantian Feng and Shrikanth Narayanan. 2024.
\newblock Foundation model assisted automatic speech emotion recognition: Transcribing, annotating, and augmenting.
\newblock In \emph{ICASSP 2024-2024 IEEE International Conference on Acoustics, Speech and Signal Processing (ICASSP)}, pages 12116--12120. IEEE.

\bibitem[{Garten et~al.(2019)Garten, Kennedy, Hoover, Sagae, and Dehghani}]{gartenIncorporatingDemographicEmbeddings2019}
Justin Garten, Brendan Kennedy, Joe Hoover, Kenji Sagae, and Morteza Dehghani. 2019.
\newblock Incorporating demographic embeddings into language understanding.
\newblock \emph{Cognitive science}, 43(1):e12701.

\bibitem[{Golazizian et~al.(2024)Golazizian, Omrani, Ziabari, and Dehghani}]{golazizian2024cost}
Preni Golazizian, Ali Omrani, Alireza~S Ziabari, and Morteza Dehghani. 2024.
\newblock Cost-efficient subjective task annotation and modeling through few-shot annotator adaptation.
\newblock \emph{arXiv preprint arXiv:2402.14101}.

\bibitem[{Gordon et~al.(2022)Gordon, Lam, Park, Patel, Hancock, Hashimoto, and Bernstein}]{gordonJuryLearningIntegrating2022}
Mitchell~L Gordon, Michelle~S Lam, Joon~Sung Park, Kayur Patel, Jeff Hancock, Tatsunori Hashimoto, and Michael~S Bernstein. 2022.
\newblock Jury learning: {{Integrating}} dissenting voices into machine learning models.
\newblock In \emph{Proceedings of the 2022 {{CHI Conference}} on {{Human Factors}} in {{Computing Systems}}}, pages 1--19.

\bibitem[{Guo et~al.(2025)Guo, Yang, Zhang, Song, Zhang, Xu, Zhu, Ma, Wang, Bi et~al.}]{guo2025deepseek}
Daya Guo, Dejian Yang, Haowei Zhang, Junxiao Song, Ruoyu Zhang, Runxin Xu, Qihao Zhu, Shirong Ma, Peiyi Wang, Xiao Bi, et~al. 2025.
\newblock Deepseek-r1: Incentivizing reasoning capability in llms via reinforcement learning.
\newblock \emph{arXiv preprint arXiv:2501.12948}.

\bibitem[{Haimson et~al.(2021)Haimson, Delmonaco, Nie, and Wegner}]{haimson2021disproportionate}
Oliver~L Haimson, Daniel Delmonaco, Peipei Nie, and Andrea Wegner. 2021.
\newblock Disproportionate removals and differing content moderation experiences for conservative, transgender, and black social media users: Marginalization and moderation gray areas.
\newblock \emph{Proceedings of the ACM on Human-Computer Interaction}, 5(CSCW2):1--35.

\bibitem[{Hartmann et~al.(2023)Hartmann, Schwenzow, and Witte}]{hartmann2023political}
Jochen Hartmann, Jasper Schwenzow, and Maximilian Witte. 2023.
\newblock The political ideology of conversational ai: Converging evidence on chatgpt’s pro-environmental, left-libertarian orientation.
\newblock \emph{Left-Libertarian Orientation (January 1, 2023)}.

\bibitem[{Hovy et~al.(2013)Hovy, Berg-Kirkpatrick, Vaswani, and Hovy}]{hovy2013learning}
Dirk Hovy, Taylor Berg-Kirkpatrick, Ashish Vaswani, and Eduard Hovy. 2013.
\newblock Learning whom to trust with {{MACE}}.
\newblock In \emph{Proceedings of the 2013 Conference of the North American Chapter of the Association for Computational Linguistics: {{Human}} Language Technologies}, pages 1120--1130.

\bibitem[{Hovy et~al.(2001)Hovy, Gerber, Hermjakob, Lin, and Ravichandran}]{hovy-etal-2001-toward}
Eduard Hovy, Laurie Gerber, Ulf Hermjakob, Chin-Yew Lin, and Deepak Ravichandran. 2001.
\newblock \href {https://www.aclweb.org/anthology/H01-1069} {Toward semantics-based answer pinpointing}.
\newblock In \emph{Proceedings of the First International Conference on Human Language Technology Research}.

\bibitem[{Jacovi and Goldberg(2020)}]{jacovi2020towards}
Alon Jacovi and Yoav Goldberg. 2020.
\newblock Towards faithfully interpretable nlp systems: How should we define and evaluate faithfulness?
\newblock In \emph{Proceedings of the 58th Annual Meeting of the Association for Computational Linguistics}, pages 4198--4205.

\bibitem[{Li and Roth(2002)}]{li-roth-2002-learning}
Xin Li and Dan Roth. 2002.
\newblock \href {https://www.aclweb.org/anthology/C02-1150} {Learning question classifiers}.
\newblock In \emph{{COLING} 2002: The 19th International Conference on Computational Linguistics}.

\bibitem[{Liu et~al.(2023)Liu, Wu, Michael, Suhr, West, Koller, Swayamdipta, Smith, and Choi}]{liu2023we}
Alisa Liu, Zhaofeng Wu, Julian Michael, Alane Suhr, Peter West, Alexander Koller, Swabha Swayamdipta, Noah~A Smith, and Yejin Choi. 2023.
\newblock We're afraid language models aren't modeling ambiguity.
\newblock In \emph{The 2023 Conference on Empirical Methods in Natural Language Processing}.

\bibitem[{Marchisio et~al.(2024)Marchisio, Dash, Chen, Aumiller, {\"U}st{\"u}n, Hooker, and Ruder}]{marchisio2024does}
Kelly Marchisio, Saurabh Dash, Hongyu Chen, Dennis Aumiller, Ahmet {\"U}st{\"u}n, Sara Hooker, and Sebastian Ruder. 2024.
\newblock How does quantization affect multilingual llms?
\newblock \emph{arXiv preprint arXiv:2407.03211}.

\bibitem[{Min et~al.(2022)Min, Lyu, Holtzman, Artetxe, Lewis, Hajishirzi, and Zettlemoyer}]{min2022rethinking}
Sewon Min, Xinxi Lyu, Ari Holtzman, Mikel Artetxe, Mike Lewis, Hannaneh Hajishirzi, and Luke Zettlemoyer. 2022.
\newblock Rethinking the role of demonstrations: {{What}} makes in-context learning work?
\newblock In \emph{Proceedings of the 2022 Conference on Empirical Methods in Natural Language Processing}, pages 11048--11064.

\bibitem[{Mohammad et~al.(2018)Mohammad, Bravo-Marquez, Salameh, and Kiritchenko}]{mohammad2018semeval}
Saif Mohammad, Felipe Bravo-Marquez, Mohammad Salameh, and Svetlana Kiritchenko. 2018.
\newblock Semeval-2018 task 1: {{Affect}} in tweets.
\newblock In \emph{Proceedings of the 12th International Workshop on Semantic Evaluation}, pages 1--17.

\bibitem[{Mokhberian et~al.(2022)Mokhberian, Hopp, Harandizadeh, Morstatter, and Lerman}]{mokhberian2022noise}
Negar Mokhberian, Frederic~R Hopp, Bahareh Harandizadeh, Fred Morstatter, and Kristina Lerman. 2022.
\newblock Noise audits improve moral foundation classification.
\newblock In \emph{2022 IEEE/ACM International Conference on Advances in Social Networks Analysis and Mining (ASONAM)}, pages 147--154. IEEE.

\bibitem[{Mokhberian et~al.(2023)Mokhberian, Marmarelis, Hopp, Basile, Morstatter, and Lerman}]{mokhberian2023capturing}
Negar Mokhberian, Myrl~G Marmarelis, Frederic~R Hopp, Valerio Basile, Fred Morstatter, and Kristina Lerman. 2023.
\newblock Capturing perspectives of crowdsourced annotators in subjective learning tasks.
\newblock \emph{arXiv preprint arXiv:2311.09743}.

\bibitem[{Ouyang et~al.(2022)Ouyang, Wu, Jiang, Almeida, Wainwright, Mishkin, Zhang, Agarwal, Slama, Ray et~al.}]{ouyang2022training}
Long Ouyang, Jeffrey Wu, Xu~Jiang, Diogo Almeida, Carroll Wainwright, Pamela Mishkin, Chong Zhang, Sandhini Agarwal, Katarina Slama, Alex Ray, et~al. 2022.
\newblock Training language models to follow instructions with human feedback.
\newblock \emph{Advances in neural information processing systems}, 35:27730--27744.

\bibitem[{Paletz et~al.(2023)Paletz, Golonka, Pandža, Stanton, Ryan, Adams, Rytting, Murauskaite, Buntain, Johns et~al.}]{paletz2023social}
Susannah~BF Paletz, Ewa~M Golonka, Nick~B Pandža, Grace Stanton, David Ryan, Nikki Adams, C~Anton Rytting, Egle~E Murauskaite, Cody Buntain, Michael~A Johns, et~al. 2023.
\newblock Social media emotions annotation guide ({{SMEmo}}): {{Development}} and initial validity.
\newblock \emph{Behavior Research Methods}, pages 1--51.

\bibitem[{Resnick et~al.(2021)Resnick, Kong, Schoenebeck, and Weninger}]{resnick2021survey}
Paul Resnick, Yuqing Kong, Grant Schoenebeck, and Tim Weninger. 2021.
\newblock Survey equivalence: A procedure for measuring classifier accuracy against human labels.
\newblock \emph{arXiv preprint arXiv:2106.01254}.

\bibitem[{Sap et~al.(2019)Sap, Card, Gabriel, Choi, and Smith}]{sap2019risk}
Maarten Sap, Dallas Card, Saadia Gabriel, Yejin Choi, and Noah~A Smith. 2019.
\newblock The risk of racial bias in hate speech detection.
\newblock In \emph{Proceedings of the 57th annual meeting of the association for computational linguistics}, pages 1668--1678.

\bibitem[{Sap et~al.(2022)Sap, Swayamdipta, Vianna, Zhou, Choi, and Smith}]{sapAnnotatorsAttitudesHow2022}
Maarten Sap, Swabha Swayamdipta, Laura Vianna, Xuhui Zhou, Yejin Choi, and Noah~A. Smith. 2022.
\newblock \href {https://doi.org/10.18653/v1/2022.naacl-main.431} {Annotators with {{Attitudes}}: {{How Annotator Beliefs And Identities Bias Toxic Language Detection}}}.
\newblock In \emph{Proceedings of the 2022 {{Conference}} of the {{North American Chapter}} of the {{Association}} for {{Computational Linguistics}}: {{Human Language Technologies}}}, pages 5884--5906. Association for Computational Linguistics.

\bibitem[{Shi et~al.(2024)Shi, Han, Lewis, Tsvetkov, Zettlemoyer, and Yih}]{shi2024trusting}
Weijia Shi, Xiaochuang Han, Mike Lewis, Yulia Tsvetkov, Luke Zettlemoyer, and Wen-tau Yih. 2024.
\newblock Trusting your evidence: Hallucinate less with context-aware decoding.
\newblock In \emph{Proceedings of the 2024 Conference of the North American Chapter of the Association for Computational Linguistics: Human Language Technologies (Volume 2: Short Papers)}, pages 783--791.

\bibitem[{Stanton et~al.(2021)Stanton, Izmailov, Kirichenko, Alemi, and Wilson}]{stanton2021does}
Samuel Stanton, Pavel Izmailov, Polina Kirichenko, Alexander~A Alemi, and Andrew~G Wilson. 2021.
\newblock Does knowledge distillation really work?
\newblock \emph{Advances in Neural Information Processing Systems}, 34:6906--6919.

\bibitem[{Swayamdipta et~al.(2020)Swayamdipta, Schwartz, Lourie, Wang, Hajishirzi, Smith, and Choi}]{swayamdipta2020dataset}
Swabha Swayamdipta, Roy Schwartz, Nicholas Lourie, Yizhong Wang, Hannaneh Hajishirzi, Noah~A Smith, and Yejin Choi. 2020.
\newblock Dataset cartography: Mapping and diagnosing datasets with training dynamics.
\newblock In \emph{Proceedings of the 2020 Conference on Empirical Methods in Natural Language Processing (EMNLP)}, pages 9275--9293.

\bibitem[{Trager et~al.(2022)Trager, Ziabari, Davani, Golazizian, Karimi-Malekabadi, Omrani, Li, Kennedy, Reimer, Reyes et~al.}]{trager2022moral}
Jackson Trager, Alireza~S Ziabari, Aida~Mostafazadeh Davani, Preni Golazizian, Farzan Karimi-Malekabadi, Ali Omrani, Zhihe Li, Brendan Kennedy, Nils~Karl Reimer, Melissa Reyes, et~al. 2022.
\newblock The moral foundations reddit corpus.
\newblock \emph{arXiv preprint arXiv:2208.05545}.

\bibitem[{Turpin et~al.(2024)Turpin, Michael, Perez, and Bowman}]{turpin2024language}
Miles Turpin, Julian Michael, Ethan Perez, and Samuel Bowman. 2024.
\newblock Language models don't always say what they think: Unfaithful explanations in chain-of-thought prompting.
\newblock \emph{Advances in Neural Information Processing Systems}, 36.

\bibitem[{Ustalov et~al.(2021)Ustalov, Pavlichenko, and Tseitlin}]{ustalov2021learning}
Dmitry Ustalov, Nikita Pavlichenko, and Boris Tseitlin. 2021.
\newblock Learning from crowds with crowd-kit.
\newblock \emph{arXiv preprint arXiv:2109.08584}.

\bibitem[{Vaswani et~al.(2017)Vaswani, Shazeer, Parmar, Uszkoreit, Jones, Gomez, Kaiser, and Polosukhin}]{vaswaniAttentionAllYou2017}
Ashish Vaswani, Noam Shazeer, Niki Parmar, Jakob Uszkoreit, Llion Jones, Aidan~N Gomez, \textbackslash~Lukasz Kaiser, and Illia Polosukhin. 2017.
\newblock Attention is all you need.
\newblock \emph{Advances in neural information processing systems}, 30.

\bibitem[{Wolf et~al.(2020)Wolf, Debut, Sanh, Chaumond, Delangue, Moi, Cistac, Rault, Louf, Funtowicz, Davison, Shleifer, von Platen, Ma, Jernite, Plu, Xu, Scao, Gugger, Drame, Lhoest, and Rush}]{wolf-etal-2020-transformers}
Thomas Wolf, Lysandre Debut, Victor Sanh, Julien Chaumond, Clement Delangue, Anthony Moi, Pierric Cistac, Tim Rault, Rémi Louf, Morgan Funtowicz, Joe Davison, Sam Shleifer, Patrick von Platen, Clara Ma, Yacine Jernite, Julien Plu, Canwen Xu, Teven~Le Scao, Sylvain Gugger, Mariama Drame, Quentin Lhoest, and Alexander~M. Rush. 2020.
\newblock \href {https://www.aclweb.org/anthology/2020.emnlp-demos.6} {Transformers: State-of-the-art natural language processing}.
\newblock In \emph{Proceedings of the 2020 Conference on Empirical Methods in Natural Language Processing: System Demonstrations}, pages 38--45, Online. Association for Computational Linguistics.

\end{thebibliography}

\clearpage


\appendix

\section{Methodology} \label{sec:appendix-method}

In this section, we present a mathematical formulation for the baseline and \liahr to avoid any potential ambiguities arising from the natural language description of the main text. We follow the notation of \citet{chochlakis2025aggregation}. For a set of examples $\mathcal{X}$, and a set of labels $\mathcal{Y}$, a dataset $\mathcal{D}^a$ defines a mapping $f^a:\mathcal{X} \rightarrow \mathcal{Y}$, where $a$ denotes a specific annotator or the aggregate. Similarly, $\mathcal{D}^a = \{(x, y)\ |\ x\in\mathcal{X}, y = f^a(x)\}$, which is characterized by joint distribution $p^a(x, y)$. The \textit{gold} query pair is denoted as $(x_q, y^a_q),\ y^a_q = f^a(x_q)$.

\paragraph{Reasonableness Baseline}

We want to sample $k$ train documents from $\mathcal{X}$ to create a prompt with document-label pairs, as well as corresponding binary reasonableness labels, denoted simply as $1$ and $0$\footnote{in the actual prompt, the labels are presented as ``yes'' or ``no'', or ``reasonable'' and ``unreasonable'', not as $1$ and $0$, as shown in Table~\ref{tab:prompt-example}.}. We choose half ($\frac{k}{2}$) pairs to have the ``reasonable'' label, and for other half the ``unreasonable'' label. To sample reasonable pairs for our prompt, we sample document-label pairs directly from $\mathcal{D}^a$ as $S^r = \{(x_i, y_i, 1):\ (x_i, y_i)\sim p^a,\ i\in[\frac{k}{2}] \}$. For unreasonable pairs, we sample the documents $x$ and the labels $y$ independently of each other from the dataset as $S^u = \{(x_i, y_i, 0):\ (x_i, y_i)\sim p^a_I,\ i\in[\frac{k}{2}] \}$, where $p^a_I(x, y) = p^a(x)p^a(y)$, in effect assigning random yet in-distribution labels to each document. The complete demonstrations for the model are $S = S^r \cup S^u$, and the order of the examples in the prompt is determined randomly. The query document is presented with the gold label $y_d^a = y^a_q$ (like $S^r$; ``Gold pair'' in the baseline figures like Figure~\ref{fig:semeval-baseline-results}) or a random label $y_d^a$ independently from the query $x_q$ using $p^a(y)$ (like $S^u$; ``Rand pair'' in the baseline figures) at the end, and we elicit the final prediction from the model, namely $1$ or $0$: $S^\prime = S \cup \{(x_q, y_d^a, \_)\}$. We show the rate of $1$ predictions in our results.

\paragraph{\liahr}

To create the prompt, we sample $k$ demonstrations for the prompt from $\mathcal{D}^a$ with $p^a$ as $S = \{(x_i, y_i):\ (x_i, y_i)\sim p^a,\ i\in[k] \}$. In \liahr, the first demonstration is the query $x_q$. For this demonstration, we either use its gold label $y_d^a = y_q^a$ (``Query w/ gold'' in the \liahr figures like Figure~\ref{fig:semeval-copy-paste-results}) or sample a random label $y_d^a$ independently from the query $x_q$ using $p^a(y)$ (``Query w/ rand'' in the \liahr figures). This query pair is included in the prompt as the first demonstration, and the query document $x_q$ is appended in the prompt, eliciting a prediction for it from the model, $S^\prime = \{(x_q, y_d^a)\} \cup S \cup \{(x_q, \_ )\}$. We measure and show the similarity between the predictions of the model with $y_d^a$, and only measure the similarity of the prediction with $y_q^a$ for ``Gold perf w/ rand'' in \liahr figures.

\section{More Implementation Details} \label{sec:appendix-more-impl}

We used A100 NVIDIA GPUs with 80GB VRAM for 70B models, and A40 NVIDIA GPUs for smaller models. The budget for OpenAI API calls was less than \$50.

For all datasets, we evaluate LLMs on the dev set. For QueerReclaimLex, we only maintain the labels with agreement between the two annotators. Our splits in the dataset were random. The evaluation set was balanced, containing 84 examples.

For the baseline, we sample the random labels for the pair similarly to the random labels in \liahr. In the demonstrations, we use equal amounts of pairs with gold labels and random labels. For Demux, we use the same training regime as \citet{chochlakisLeveragingLabelCorrelations2023}, using the intra loss with a coefficient of $0.2$, but training only on the train set instead of integrating the dev set in training after early stopping. Confidence intervals are shown from 3 model runs.

We present examples of all the prompts in Table~\ref{tab:prompt-example}. Across each dataset, the same examples are used in the prompt within each seed when the number of shots is equal. That means that, for example, in QueerReclaimLex, the only factor that is different across the four settings presented in Figure~\ref{fig:diversity-queer} is the causally controlled labels reflecting in-group or out-group perspectives.

\begin{table}[!t]
    \centering
    \footnotesize
    \renewcommand{\arraystretch}{1}
    \begin{tabular}[t]{@{}p{23em} @{}}

    \textbf{SemEval 2018 Task 1 E-c} \\
    \narrowbotc{Classify the following inputs into none, one, or multiple the following emotions per input: anger, anticipation, disgust, fear, joy, love, optimism, pessimism, sadness, surprise and trust.\\\\Input:  `Skipping meals is the national way of showing disappointment and . wiz`\\\{"label": ["anger", "disgust", "sadness"]\}\\\\Input: `But guess what ? I'm sober`} \\ \\

    \textbf{GoEmotions}  \\
    
    \narrowbotc{Classify the following inputs into none, one, or multiple the following emotions per input: joy, optimism, admiration, surprise, fear, sadness and anger.\\\\Input: `Is this in New Orleans?? I really feel like this is New Orleans.`\\\{"label": []\}\\\\Input: `BANG motherfucker!`} \\\\

    \textbf{MFRC}  \\

    \narrowbotc{Classify the following inputs into none, one, or multiple the following moral foundations per input: authority, care, equality, loyalty, proportionality and purity.\\\\ Input: `Le Pen was absolutely terrible. That's pretty much all we can take out of this debate.`\\\{"label": []\}\\\\Input: `It’s just capitalism is a pure distillation of those bad traits`}\\\\

    \textbf{QueerReclaimLex} \\

    \narrowbotc{Consider whether the following inputs present harm or not to the mentioned LGBTQ group, and answer with: no harm and harm.\\\\Input: `fags ew`\\Assessment: no harm\\\\Input: `went from edm junkie to college dyke way too quick`\\Assessment:}\\\\

    \textbf{Reasonableness baseline} \\

    \narrowbotc{Assess the reasonableness of the provided label for each input. Namely, evaluate whether the label makes sense for its corresponding input, under some reasonable interpretation. Reply only with unreasonable and reasonable.\\\\Input: `Skipping meals is the national way of showing disappointment and . wiz`\\Label: surprise, optimism\\Assessment: unreasonable\\\\Input: `But guess what ? I'm sober`\\Label: joy\\Assessment: }
    
    \end{tabular}
    \caption{Prompt template examples}
    \label{tab:prompt-example}
\end{table}

\begin{figure}[!ht]
    \centering
    \includegraphics[width=1\linewidth]{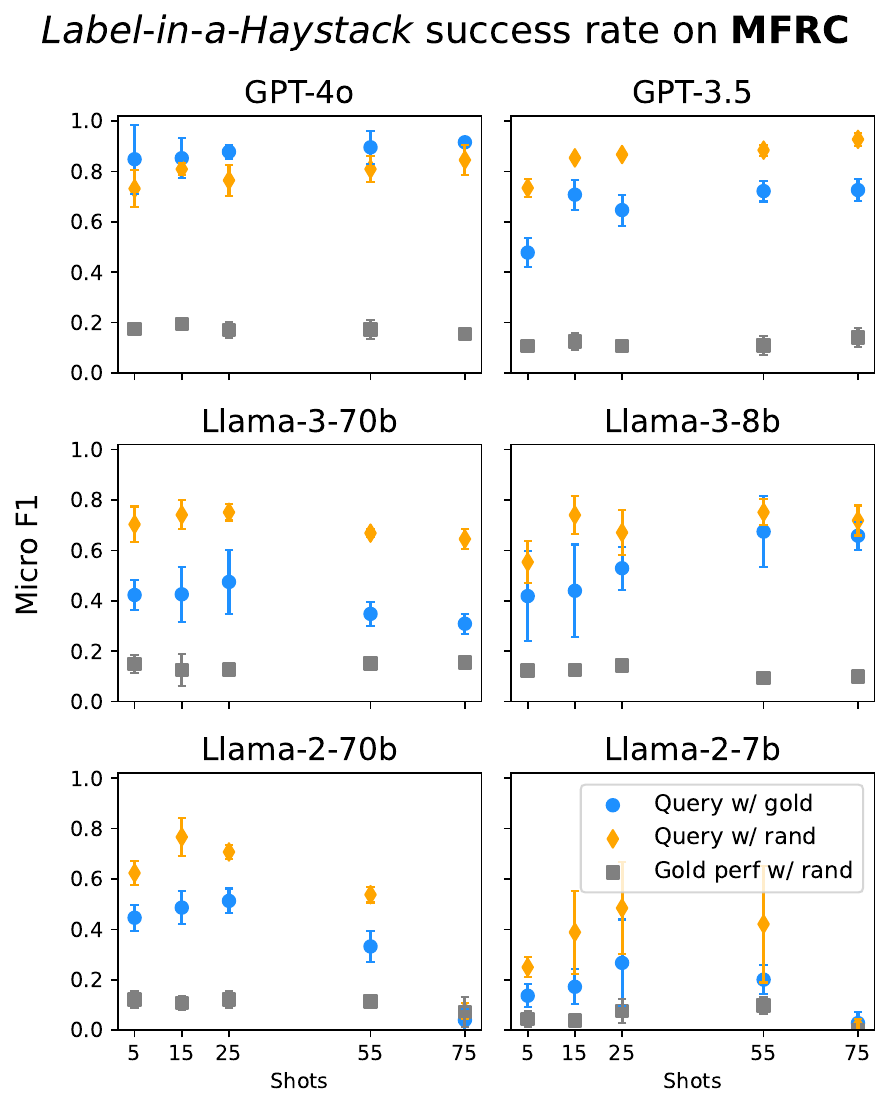}
    \caption{Success rate of copying with \liahr on \textbf{MFRC} when using the gold and random labels for the query in the prompt across various numbers of demonstrations. We also show performance w.r.t. the gold labels when using random query labels.}
    \label{fig:mfrc-copy-paste-results}
\end{figure}

\begin{figure}[!ht]
    \centering
    \includegraphics[width=1\linewidth]{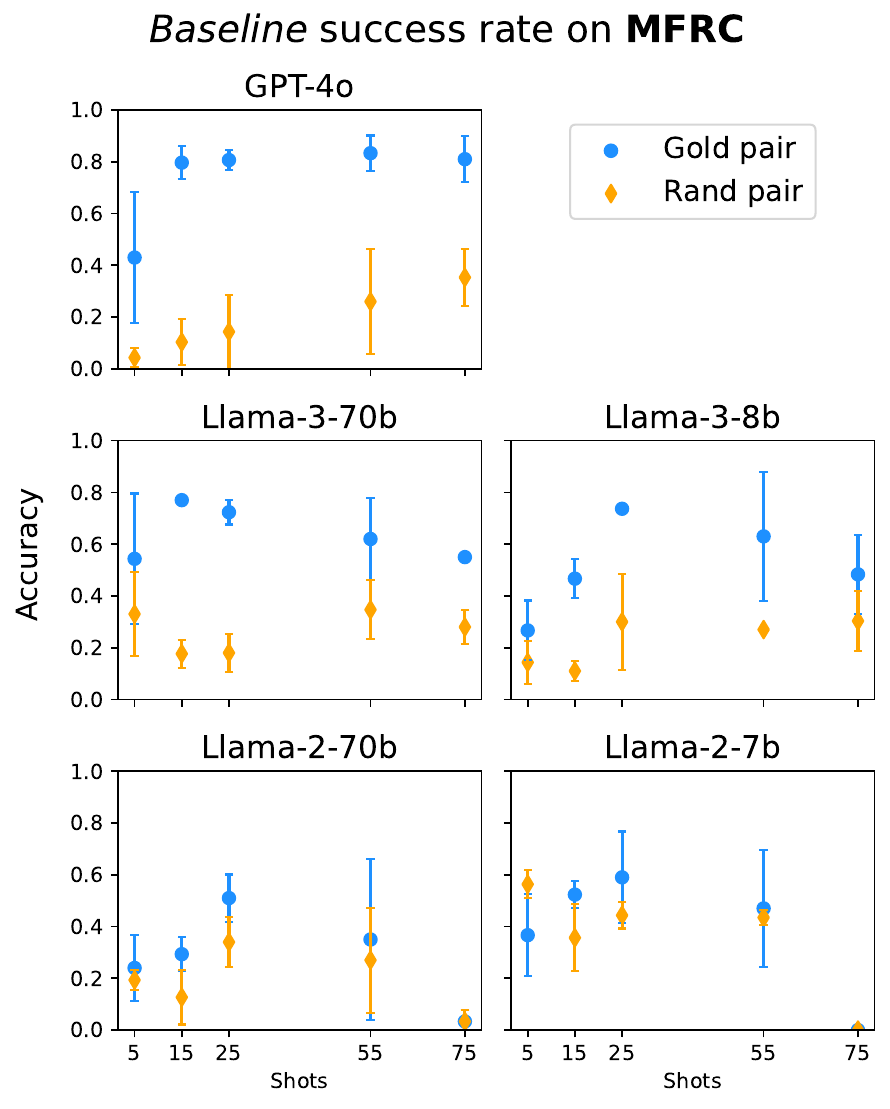}
    \caption{\textit{Baseline} ``reasonable'' scores on \textbf{MFRC} when using gold and random input-label pairs.}
    \label{fig:mfrc-baseline-results}
\end{figure}

\begin{figure*}[!ht]
    \centering
    \includegraphics[width=0.8\linewidth]{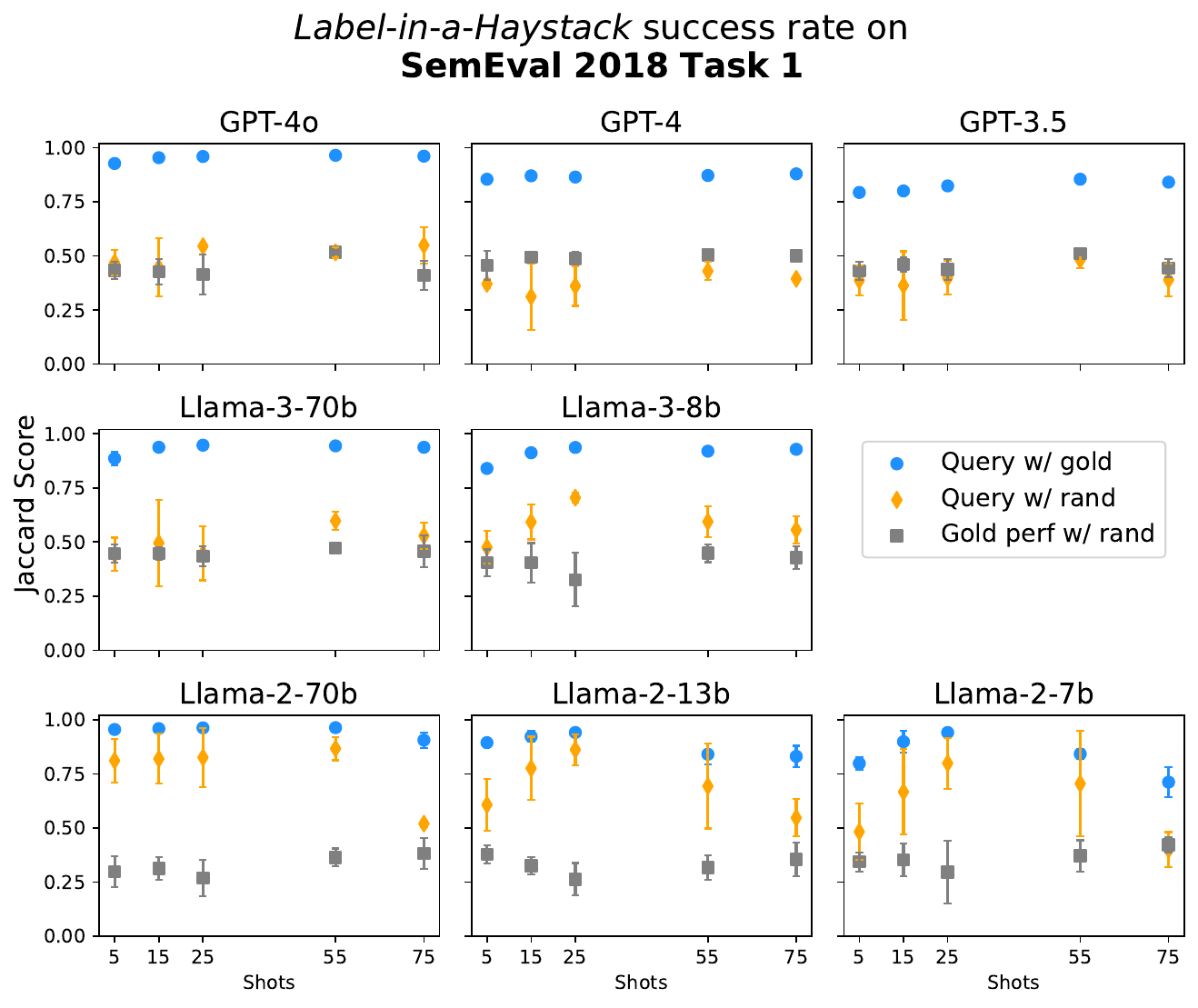}
    \caption{\textit{Full} scores on the copy-paste task on \textbf{SemEval} when using the gold and random labels for the query in the prompt across various numbers of demonstrations. We also show performance w.r.t. the gold labels when using random query labels.}
    \label{fig:semeval-deprecated}
\end{figure*}

\section{Full Human Evaluations} \label{sec:appendix-human-eval}

In human evaluations, to avoid biasing annotators towards specific answers --- for example, having the dataset label always as the first option~\cite{turpin2024language} ---, we randomly interleave reasonable and unreasonable examples (label according to the model) for the \textbf{Reasonableness} check, whereas for the \textbf{Preference} check, we randomly change the order with which the label in the dataset and the model's prediction are presented to the annotators. We present full results, including the number of trials and the precise numbers of them, here in Table~\ref{tab:full-significance} and Table~\ref{tab:full-significance-baseline}. We note that we recruited 11 colleagues (6 males and 5 females, ages 20-28, students or researchers) to annotate to get as many perspectives as possible and avoid biasing the result. Note that the annotators were shown the \textit{Reasonableness baseline} prompt from Table~\ref{tab:prompt-example}, modified appropriately.

\begin{table*}[!ht]
    \centering
    \begin{tabular}{cl ccc cc}
        & & \multicolumn{3}{c}{\textbf{Reasonableness}} & \multicolumn{2}{c}{\textbf{Preference}} \\
        \cmidrule(lr){3-5} \cmidrule(lr){6-7}
        && Correct Ratio & Wrong Ratio & p-value & Ratio & p-value \\
        \cmidrule{2-7}
        \parbox[t]{5mm}{\multirow{9}{*}{\rotatebox[origin=c]{90}{\textbf{SemEval}}}}
        &\hspace{-8px} \liahr \\
        &Llama-3 70b & 31/14 & 28/17 & 6.57e-1 & 26/12 & 3.36e-2 \\
        &GPT-3.5 & 27/8 & 29/16 & 9.52e-2 & 54/36 & 7.25e-2 \\
        &GPT-4 & 25/5 & 4/26 & 2.38e-7 & 41/15 & 6.86e-4   \\
        &GPT-4o & 60/20 & 21/31 & 1.40e-4 & 49/16 & 5.08e-5 \\\\

        &\hspace{-8px}\textit{baseline} \\
        &Llama-3 70b & 48/12 & 29/31 & 6.11e-4 & - & - \\
        &GPT-4o & 90/10 & 49/51 & 8.08e-10 & - & - \\
        
        \cmidrule{2-7}
        \parbox[t]{5mm}{\multirow{5}{*}{\rotatebox[origin=c]{90}{\textbf{GoEmotions}}}}
        &\hspace{-8px}\liahr \\
        &GPT-4o & 43/14 & 9/27 & 5.12e-6 & 36/3 & 3.61e-8 \\
        \\
        &\hspace{-8px}\textit{baseline} \\
        &GPT-4o & 57/23 & 33/47 & 2.47e-4 & - & - \\
        
    \end{tabular}
    \caption{Results of statistical ananlysis for \liahr on \textbf{SemEval} and \textbf{GoEmotions}. \textbf{Correct Ratio} refers to proportion of dataset labels deemed reasonable vs. unreasonable by annotators when the model performed the copy-paste task correctly, and similarly for \textbf{Wrong Ratio} when the copy-paste task was performed incorrectly. \textbf{Ratio} reflects the times the model's labels were preferred over the gold labels (when the model performed copy-pasting incorrectly).}
    \label{tab:full-significance}
\end{table*}

\begin{table}[!ht]
    \centering
    \begin{tabular}{l cc}
        \multirow{2}{*}{\textbf{Model}} & \multicolumn{2}{c}{\textbf{Preference}} \\
        \cmidrule(lr){2-3}
        & Ratio & p-value \\
        \midrule
        GPT-3.5 & 33/27 & 0.519 \\
        GPT-4 & 28/32 & 1 \\
    \end{tabular}
    \caption{Results of statistical analysis for the regular ICL / raw predictions setting on \textbf{SemEval}. \textbf{Ratio} reflects the times the model's predictions were preferred over the gold labels.}
    \label{tab:full-significance-baseline}
\end{table}

\section{More models on SemEval properties} \label{sec:appendix-deprecated-semeval}

Here, we present additional results on SemEval with some deprecated models present in Figure~\ref{fig:semeval-deprecated}. We see, interestingly, that GPT-4 shows a better performance profile than GPT-4o, indicating that the models have successfully been trained to become more compliant to the user, even if the model disagrees, potentially decreasing the utility of \liahr.

\section{MFRC properties} \label{sec:appendix-mfrc-properties}

In this section, we present the results for \textbf{Nonconformity}, \textbf{Rectification}, and \textbf{Noise rejection} in MFRC, in Figures~\ref{fig:mfrc-copy-paste-results} and \ref{fig:mfrc-baseline-results}.

We observe that even GPT-3.5 does not achieve \textbf{Noise Rejection} and \textbf{Rectification}, but GPT-4o is showing positive trends in the criteria we have. Interestingly, there seem to be settings were random labels perform better than the gold ones. Here, we hypothesize that this happens because we always sample at least one label for the random label case, whereas the dataset contains many examples with no labels. 

\section{Results on objective tasks} \label{sec:appendix-objective}

\begin{figure}[!ht]
    \centering
    \includegraphics[width=1\linewidth]{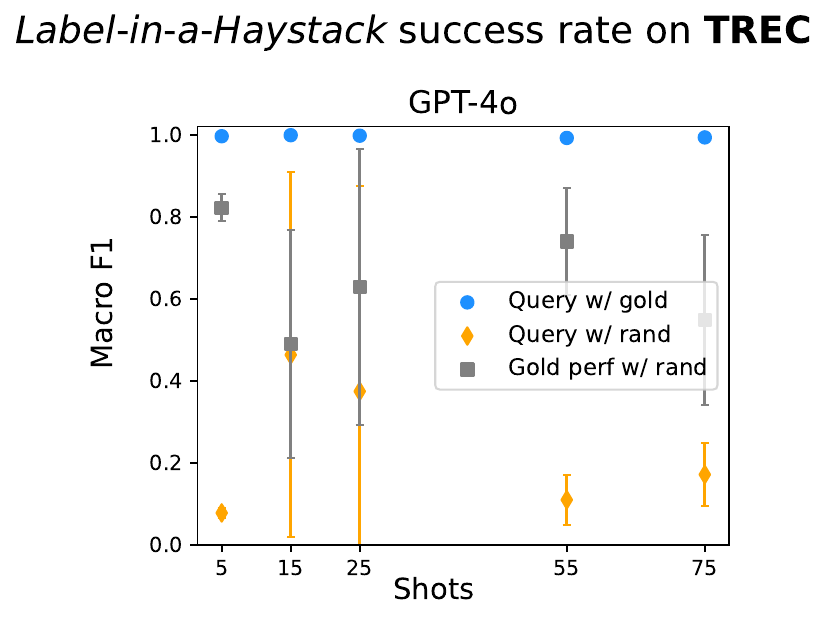}
    \caption{Scores with \liahr on \textbf{TREC} (objective benchmark) when using the gold and random labels for the query in the prompt across various numbers of demonstrations. We also show performance w.r.t. the gold labels when using random query labels.}
    \label{fig:trec}
\end{figure}

Here, we present some experimental results on an objective task, the \textbf{T}ext \textbf{RE}trieval \textbf{C}onference (\textbf{TREC}) question classification benchmark~\cite{li-roth-2002-learning, hovy-etal-2001-toward}, which contains annotations for the type of information the question pertains to, and specifically Abbreviation, Entity, Description and abstract concept, Human being, Location, and Numeric value. We show these results to verify the intuition that, in principle, \liahr can be used for objective tasks too. Indeed, we see in Figure~\ref{fig:trec}, the system meets our defined properties, with the \textbf{Rectification} being, in fact, very strong in this objective setting, suggesting the models in some ways, at least implicitly, learn to represent the nuanced difference between objective and subjective tasks.

\section{Degradation in copy-paste performance} \label{sec:appendix-degradation}

\begin{figure}[!ht]
    \centering
    \includegraphics[width=1\linewidth]{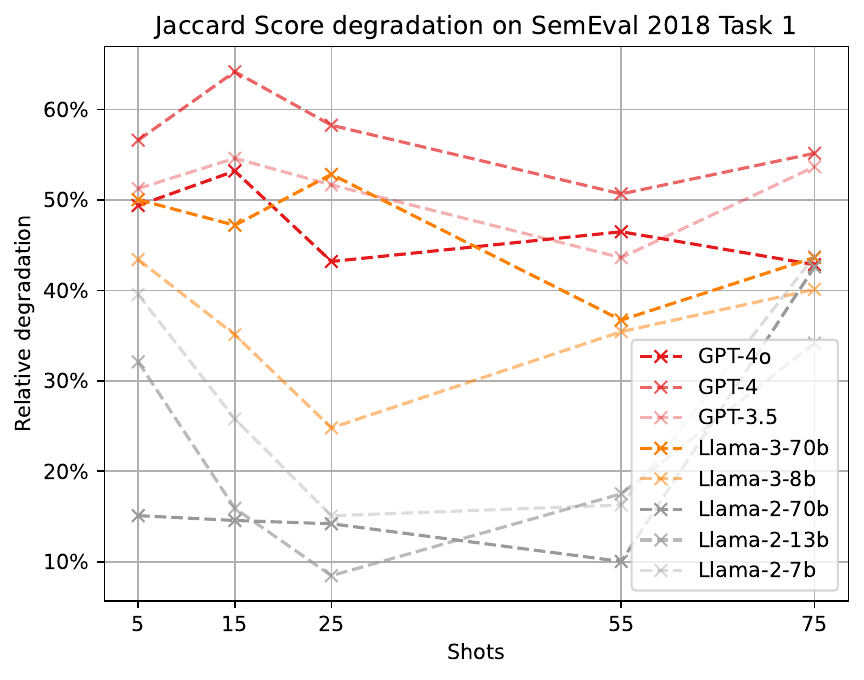}
    \caption{Degradation in copy-paste performance on \textbf{SemEval} when using random labels compared to the dataset's labels.}
    \label{fig:semeval-degradation}
\end{figure}

\begin{figure}[!ht]
    \centering
    \includegraphics[width=1\linewidth]{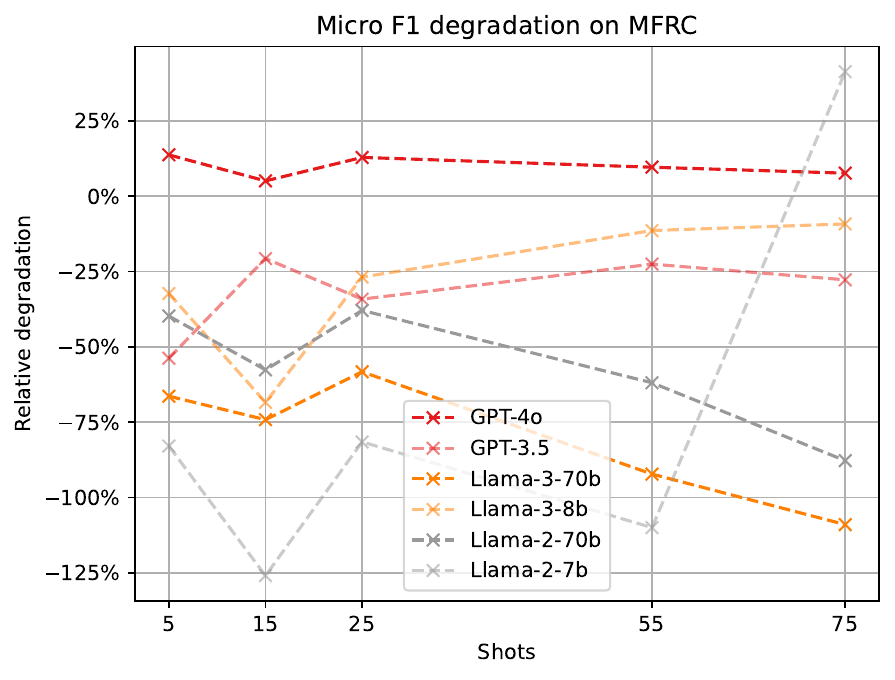}
    \caption{Degradation in copy-paste performance on \textbf{MFRC} when using random labels compared to the dataset's labels.}
    \label{fig:mfrc-degradation}
\end{figure}

\begin{figure}[!ht]
    \centering
    \includegraphics[width=1\linewidth]{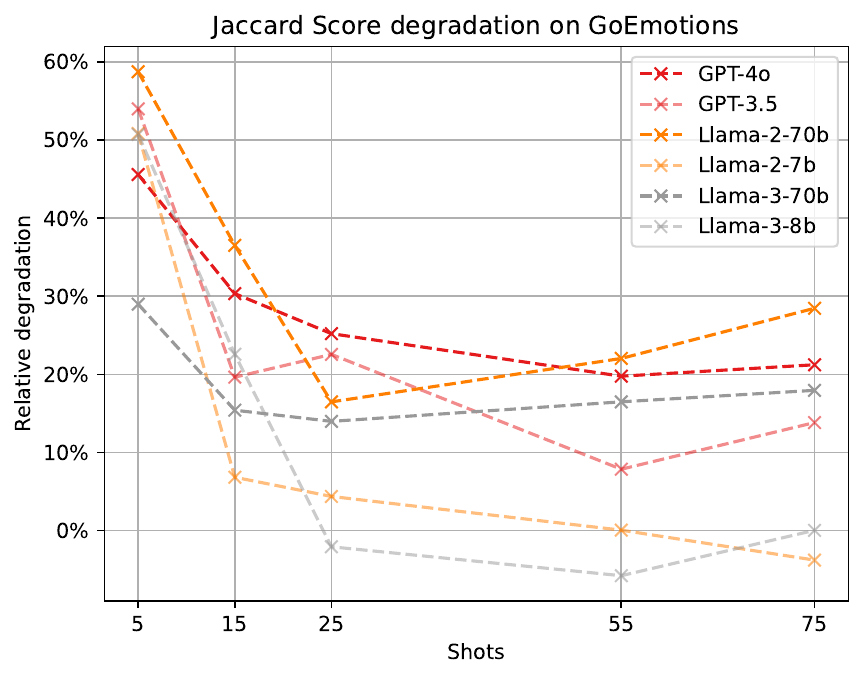}
    \caption{Degradation in copy-paste performance on \textbf{GoEmotions} when using random labels compared to the dataset's labels.}
    \label{fig:goemotions-degradation}
\end{figure}

In this section, as a summary of our results, we present how different model families and scale affects the drop in copy-paste performance when switching from the gold label for the demo query to a random label in \textit{Label in a Haystack}. We demonstrate the results for SemEval in Figure~\ref{fig:semeval-degradation}, for GoEmotions in Figure~\ref{fig:goemotions-degradation}, and MFRC in Figure~\ref{fig:mfrc-degradation}. It is interesting to look at the three model families and observe that the more capable the model family is, the larger the degradation in performance tends to be. Moreover, within each family, the larger models usually end up with worse degradation, except for the least capable Llama-2 in some instances, where the trend is the opposite. We therefore hypothesize that there is a U-shaped trend, where, on the lower end, the ability to better follow instructions leads to smaller degradations in performance when shifting to random labels. However, as models continue to get larger, the pull of the priors on the posteriors becomes greater~\cite{chochlakis2024strong}, leading to greater degradation.

\begin{figure*}[!ht]
    \centering
    \includegraphics[width=1\linewidth]{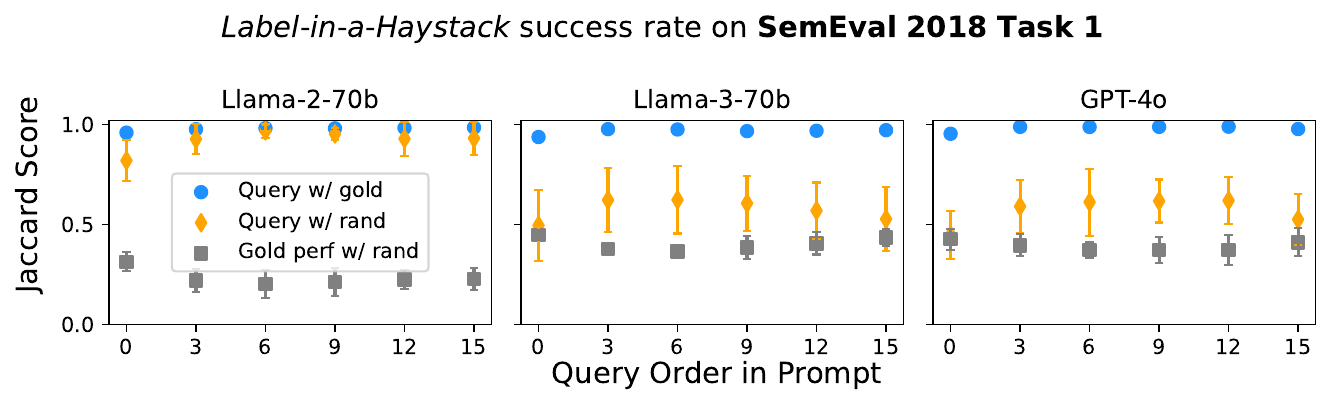}
    \caption{Scores on the 15-shot \liahr on \textbf{SemEval} when changing the position of the query in the demonstrations.}
    \label{fig:semeval-order}
\end{figure*}

\begin{table}[!ht]
    \centering
    \begin{tabular}{lcc}
        \multirow{2}{*}{\textbf{Setting}} & \multicolumn{2}{c}{\textbf{Jaccard Score}} \\
        \cmidrule(lr){2-3}
        & \textbf{GoEmotions} & \textbf{SemEval} \\
        \midrule
        \texttt{Original} & \rescell{0.623}{0.001} & \rescell{\textbf{0.574}}{0.001} \\
        \texttt{Replaced} & \rescell{\textbf{0.624}}{0.002} & \rescell{\textbf{0.574}}{0.003} \\
        \texttt{Replaced (trn)} & \rescell{0.615}{0.001} & \rescell{0.562}{0.001} \\
        \texttt{Filtered} & \rescell{\textbf{0.624}}{0.003} & \rescell{0.561}{0.002} \\
        \texttt{Bsl Filtered} & \rescell{0.615}{0.002} & \rescell{0.558}{0.002} \\
        \texttt{Predictions} & \rescell{0.430}{0.004} & \rescell{0.474}{0.000} \\
    \end{tabular}
    \caption{Performance of BERT-based \textit{Demux} on various settings using LLM label corrections.}
    \label{tab:js-setting}
\end{table}

\section{Extra Ecological Validity results} \label{sec:appendix-eco-val}

For completeness, we also present the Jaccard Score for our ecological validity studies to supplement the Micro F1 present in the main body. Results in Table~\ref{tab:js-setting} show similar as in Table~\ref{tab:microf1-setting}.

\section{Filtering per Label} \label{sec:appendix-filter-per-label}

We present the success rate of each individual label for our 3 main datasets in Table~\ref{tab:semeval-label-scores}, \ref{tab:goemotions-label-scores}, and \ref{tab:mfrc-label-scores} based on a 25-shot run with GPT-4o. We see that no label is disproportionately affected, except \textit{trust} in SemEval, the label with the least amount of annotations. On GoEmotions, scores are generally lower compared to GoEmotions, reflecting the clustering process that has been applied to shrink the label set to a reasonable amount.

\begin{table}[!ht]
    \centering
    \begin{tabular}{lc}
        \textbf{Emotion} & \textbf{F1} \\
        \midrule
         anger & \rescell{0.972}{0.016} \\
        anticipation & \rescell{0.921}{0.017} \\
        disgust & \rescell{0.939}{0.019} \\
        fear & \rescell{0.977}{0.016} \\
        joy & \rescell{0.965}{0.010} \\
        love & \rescell{0.973}{0.019} \\
        optimism & \rescell{0.995}{0.007} \\
        pessimism & \rescell{0.922}{0.034} \\
        sadness & \rescell{0.994}{0.008} \\
        surprise & \rescell{1.000}{0.000} \\
        trust & \rescell{0.867}{0.094} \\
    \end{tabular}
    \caption{Success rates of \liahr on SemEval using 25-shot GPT-4o.}
    \label{tab:semeval-label-scores}
\end{table}

\begin{table}[!ht]
    \centering
    \begin{tabular}{lc}
        \textbf{Emotion} & \textbf{F1} \\
        \midrule
        admiration & \rescell{0.950}{0.021} \\
        anger & \rescell{0.973}{0.000} \\
        fear & \rescell{1.000}{0.000} \\
        joy & \rescell{0.871}{0.020} \\
        optimism & \rescell{0.908}{0.036} \\
        sadness & \rescell{0.930}{0.028} \\
        surprise & \rescell{0.944}{0.020} \\
    \end{tabular}
    \caption{Success rates of \liahr on GoEmotions using 25-shot GPT-4o.}
    \label{tab:goemotions-label-scores}
\end{table}

\begin{table}[!ht]
    \centering
    \begin{tabular}{lc}
        \textbf{Moral foundation} & \textbf{F1} \\
        \midrule
        authority & \rescell{0.889}{0.157} \\
        care & \rescell{0.939}{0.043} \\
        equality & \rescell{0.978}{0.031} \\
        loyalty & \rescell{0.974}{0.036} \\
        proportionality & \rescell{1.000}{0.000} \\
    \end{tabular}
    \caption{Success rates of \liahr on GoEmotions using 25-shot GPT-4o.}
    \label{tab:mfrc-label-scores}
\end{table}

\section{Position of Label in the Haystack} \label{sec:appendix-position}

We also experiment with changing the position of the query in the prompt and evaluating how all our metrics change. We present our results in Figure~\ref{fig:semeval-order}, with standard deviations shown. We see that no major changes are observed in the predictions of the model, irrespective of where the query appears in the demonstrations. It is very interesting to see that even when the query is the last demonstration (just before itself then), the results remain remarkably similar to when it appears first in the prompt, separated by 15 examples with itself.

\begin{figure}[!t]
    \centering
    \includegraphics[width=1\linewidth]{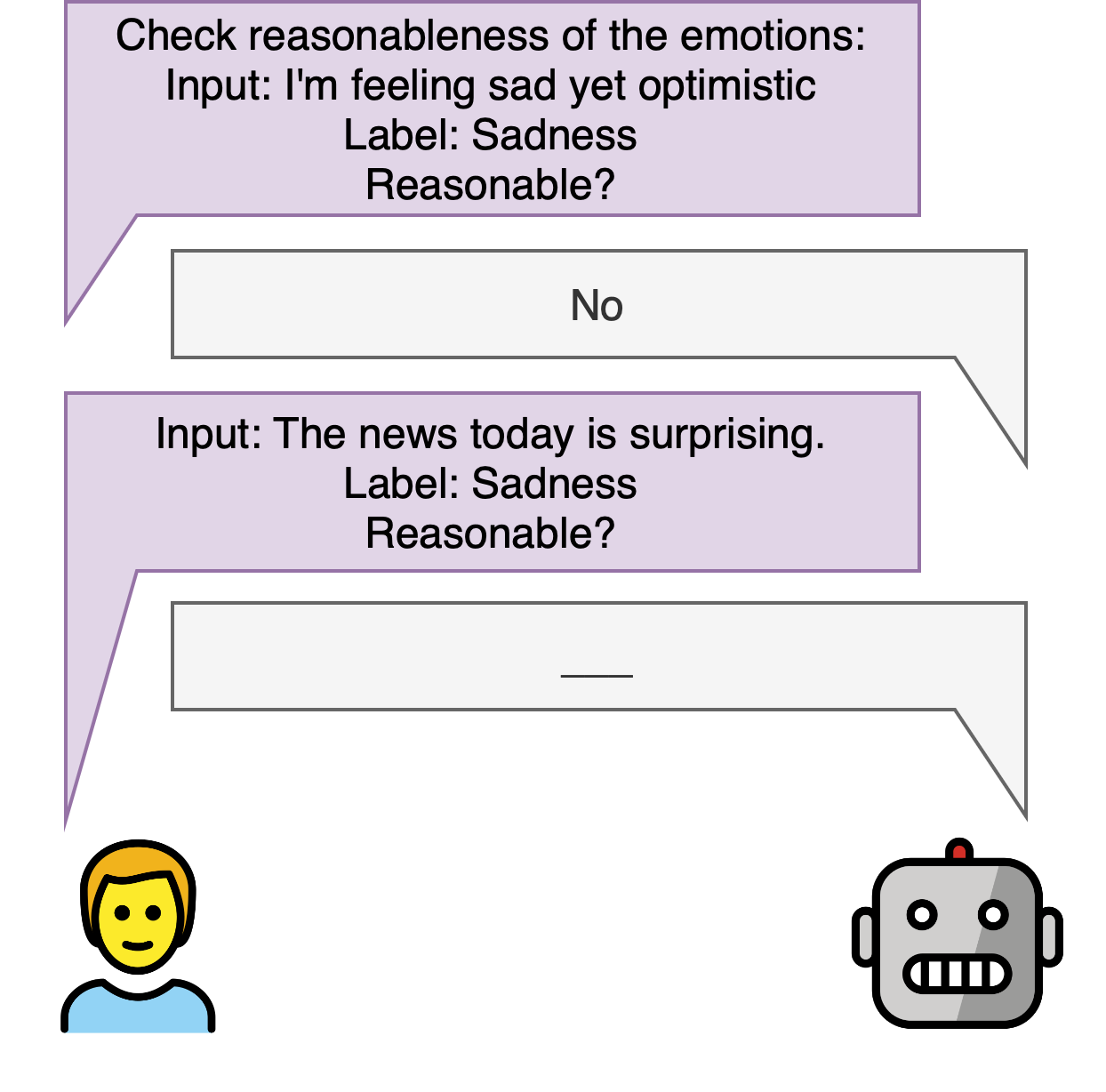}
    \caption{Reasonableness labels: The model is instructed to perform a reasonableness check, as captured by the label names. However, we check for the ability of the model to correctly copy-paste the query's label from its prompt.}
    \label{fig:reasonableness-labels}
\end{figure}

\section{Overall Reasonableness of Annotations}

We can estimate the overall reasonableness of the datasets by using our existing analyses. For example, we present the derivation process for SemEval using the GPT-4o (15-shot) \liahr results. First, looking at Figure~\ref{fig:semeval-copy-paste-results}, we can derive the percentage of human annotations predicted to be reasonable by \liahr, $p(\text{\liahr reasonable}) = 0.954$. Then, focusing on Table~\ref{tab:full-significance}, we can derive the proportion of examples annotated as unreasonable by our annotators both when \liahr predicted reasonable and unreasonable, namely $p(\text{reasonable}\ |\ \text{\liahr reasonable}) = \frac{42}{60}$,  $p(\text{reasonable}\ |\ \text{\liahr unreasonable}) = \frac{12}{39}$. Finally, we can estimate the overall reasonableness of the annotations as:
\begin{equation}
    \begin{split}
        p(\text{reasonable}) &= \\
        p(\text{reasonable}\ &|\ \text{\liahr reasonable}) \\
        &\cdot p(\text{\liahr reasonable}) \\
        + p(\text{reasonable}\ &|\ \text{\liahr unreasonable}) \\
        &\cdot (1 - p(\text{\liahr reasonable})) \\
        &= 0.682.
    \end{split}
\end{equation}
The same estimate, when checking with Llama-3 70b, comes to $0.625$, and with GPT-3.5 to $0.727$. The results are only an approximation, since \liahr results are presented in Jaccard Score, not accuracy. The same procedure can be used with the baseline, deriving more theoretically sound estimates.

\begin{table*}[!ht]
    \centering
    \begin{tabular}{lccc}
        & \textbf{SemEval} & \textbf{GoEmotions} & \textbf{MFRC} \\
        \midrule
        \multicolumn{4}{l}{\textbf{\filterbert}} \\
        \quad Gold pairs & 0.824 $\pm$ 0.017 & 0.751 $\pm$ 0.010 & 0.181 $\pm$ 0.006 \\
        \quad Rand pairs & 0.244 $\pm$ 0.010 & 0.215 $\pm$ 0.011 & 0.039 $\pm$ 0.008 \\
        \addlinespace
        \multicolumn{4}{l}{\textbf{GPT-4o (5-shot)}} \\
        \quad Gold pairs & 0.887 $\pm$ 0.078 & 0.693 $\pm$ 0.063 & 0.430 $\pm$ 0.254 \\
        \quad Rand pairs & 0.277 $\pm$ 0.235 & 0.243 $\pm$ 0.029 & 0.043 $\pm$ 0.038 \\
    \end{tabular}
    \caption{Filtering accuracy for “reasonable” vs. “unreasonable” label-document pairs using a proxy supervised BERT-based classifier (trained on all data) and GPT-4o (5-shot). Gold pairs match the true label; Rand pairs use randomly sampled labels.}
    \label{tab:bert-vs-gpt-filtering}
\end{table*}

\section{BERT-based Filtering Baseline} \label{sec:appendix-filterbert}

In this section, we present results for a BERT-based filtering baseline, \filterbert. \filterbert's output is the binary decision of whether to filter a document-label pair out of the data pool. It is a proxy supervised baseline, as we do not use actual annotated data for reasonableness, but instead use the same strategy as for the LLM baseline to construct data. Namely, documents that are paired with their gold label from the dataset are considered ``reasonable'' pairs. To create ``unreasonable'' pairs for a document, we sample the labels from a random document in the dataset. Practically, we use all pairs from the original dataset, and for each document we also create an ``unreasonable'' pair, doubling the size of the dataset. The input is formatted similarly to Demux~\cite{chochlakisLeveragingLabelCorrelations2023}, where the input consists of the \texttt{CLS} token, followed by the candidate labels, in turn followed by a \texttt{SEP} token, and finally the input document. An example input, therefore, is ``\texttt{[CLS] anger, anticipation, optimism [SEP] I DIDN'T ASK FOR THIS EITHER IT JUST HAPPENED}''. We use the contextual embedding of \texttt{CLS} with a two-layer neural network (again, similar to Demux) to make the final binary prediction with a threshold of $0.5$ on the output sigmoid function. Training details are otherwise identical to Demux (note that we have removed the intro loss coefficient because we do not apply the classifier on each emotion of the prompt). Our results are presented in Table~\ref{tab:bert-vs-gpt-filtering}, in comparison to the 5-shot GPT-4o baseline results we have already presented in Figures~\ref{fig:semeval-baseline-results}, \ref{fig:goemotions-baseline-results} and \ref{fig:mfrc-baseline-results}.

The BERT-based model cannot be used to conduct the ecological validity tests due to the fact that it itself needs to use the train and dev sets to be trained, so a more direct comparison is not possible with our current setting, a shortcoming of using a BERT-based model for filtering. However, from these existing results, GPT-4o seems to align more with our intuitions of how a model should perform. For SemEval, the performance of GPT-4o is closer to random baseline performance compared to BERT in rejecting emotions, and accepts more labels. For GoEmotions, the BERT-based model seems to learn the noise in GoEmotion arising from the hierarchical clustering that we apply, achieving higher acceptance rates (as we have mentioned before). The superiority of GPT-4o on MFRC is evident.

\end{document}